\pdfoutput=1
\documentclass[11pt]{article}

\usepackage[]{acl}

\usepackage{times}
\usepackage{latexsym}

\usepackage[T1]{fontenc}
\usepackage[utf8]{inputenc}

\usepackage{microtype}

\usepackage{url}
\usepackage{amsmath}
\usepackage{amssymb}
\usepackage{graphicx}
\usepackage{graphics}
\usepackage{booktabs}
\usepackage{multirow}
\usepackage{color}
\usepackage{fixltx2e}
\usepackage{enumitem}
\usepackage{pgfplots}
\usepackage{comment}
\usepackage{nicefrac}
\usepackage{amsfonts}
\usepackage{bm}
\usepackage{bbm}
\usepackage{algorithm}
\usepackage{algpseudocode}

\usepackage{enumitem}
\setlist[itemize,enumerate]{leftmargin=*}
\usepackage[normalem]{ulem}
\usepackage{pgfplots}
\usetikzlibrary{pgfplots.groupplots}
\pgfplotsset{compat=1.3}
\usepackage{tikz}
\usetikzlibrary{patterns}
\usepackage{xcolor}
\usepackage{todonotes}
\usepackage{listings}
\usepackage{multirow}
\usepackage{graphicx}

\usepackage{inconsolata}

\usepackage{soul}
\usepackage{CJKutf8}

\definecolor{SRC}{HTML}{A0B1BA}
\definecolor{SRCREF}{HTML}{d3f8e2}
\definecolor{REF}{HTML}{a9def9}

\colorlet{SRC}{SRC!30}
\colorlet{SRCREF}{SRCREF!50}
\colorlet{REF}{REF!50}

\definecolor{plotcolor1}{HTML}{AF58BA}
\definecolor{plotcolor2}{HTML}{00CD6C}
\definecolor{plotcolor3}{HTML}{FFC61E}
\definecolor{plotcolor4}{HTML}{009ADE}

\colorlet{plotcolor1}{plotcolor1!90}
\colorlet{plotcolor2}{plotcolor2!90}
\colorlet{plotcolor3}{plotcolor3!90}
\colorlet{plotcolor4}{plotcolor4!90}

\lstdefinestyle{yamlStyle}{
  basicstyle=\footnotesize\ttfamily,
  breaklines=false,
  frame=single,
  numbers=left,
  numberstyle=\scriptsize,
  stepnumber=1,
  numbersep=8pt,
  showstringspaces=false,
  tabsize=2,
  captionpos=b,
  keywordstyle=\color{blue},
  commentstyle=\color{gray},
  stringstyle=\color{red},
  morecomment=[l]{\#},
  morekeywords={true,false,null,y,n},
  emph={},
  emphstyle=\color{black}
}

\definecolor{fc}{HTML}{1E90FF}
\definecolor{h}{HTML}{228B22}
\definecolor{bias}{HTML}{87CEFA}
\definecolor{noise}{HTML}{8B008B}
\definecolor{conv}{HTML}{FFA500}
\definecolor{pool}{HTML}{B22222}
\definecolor{up}{HTML}{B22222}
\definecolor{view}{HTML}{FFFFFF}
\definecolor{bn}{HTML}{FFD700}
\tikzset{fc/.style={black,draw=black,fill=fc,rectangle,minimum height=1cm}}
\tikzset{h/.style={black,draw=black,fill=h,rectangle,minimum height=1cm}}
\tikzset{bias/.style={black,draw=black,fill=bias,rectangle,minimum height=1cm}}
\tikzset{noise/.style={black,draw=black,fill=noise,rectangle,minimum height=1cm}}
\tikzset{conv/.style={black,draw=black,fill=conv,rectangle,minimum height=1cm}}
\tikzset{pool/.style={black,draw=black,fill=pool,rectangle,minimum height=1cm}}
\tikzset{up/.style={black,draw=black,fill=up,rectangle,minimum height=1cm}}
\tikzset{view/.style={black,draw=black,fill=view,rectangle,minimum height=1cm}}
\tikzset{bn/.style={black,draw=black,fill=bn,rectangle,minimum height=1cm}}

\definecolor{battleshipgrey}{rgb}{0.3, 0.3, 0.3}
\definecolor{brilliantrose}{rgb}{1.0, 0.33, 0.64}
\definecolor{americanrose}{rgb}{1.0, 0.01, 0.24}
\definecolor{jweigreen}{rgb}{0,0.45,0.24}
\definecolor{bluegray}{rgb}{0.1, 0.1, 0.4}
\definecolor{ao(english)}{rgb}{0.0, 0.5, 0.0}
\definecolor{blanchedalmond}{rgb}{1.0, 0.92, 0.8}
\definecolor{atomictangerine}{rgb}{1.0, 0.6, 0.4}
\definecolor{chocolate(web)}{rgb}{0.82, 0.41, 0.12}
\definecolor{bananayellow}{rgb}{1.0, 0.88, 0.21}
\definecolor{goldenbrown}{rgb}{0.6, 0.4, 0.08}
\definecolor{aliceblue}{rgb}{0.94, 0.97, 1.0}
\definecolor{beige}{rgb}{0.96, 0.96, 0.86}
\definecolor{babyblue}{rgb}{0.54, 0.81, 0.94}
\definecolor{camel}{rgb}{0.76, 0.6, 0.42}
\definecolor{cinnamon}{rgb}{0.82, 0.41, 0.12}
\definecolor{deepskyblue}{rgb}{0.0, 0.75, 1.0}
\definecolor{frenchblue}{rgb}{0.0, 0.45, 0.73}
\definecolor{classicrose}{rgb}{0.98, 0.8, 0.91}
\definecolor{frenchrose}{rgb}{0.96, 0.29, 0.54}
\definecolor{frenchlilac}{rgb}{0.53, 0.38, 0.56}
\definecolor{frenchbeige}{rgb}{0.65, 0.48, 0.36}

\definecolor{paired-light-blue}{RGB}{198, 219, 239}
\definecolor{paired-dark-blue}{RGB}{49, 130, 188}
\definecolor{paired-light-orange}{RGB}{251, 208, 162}
\definecolor{paired-dark-orange}{RGB}{230, 85, 12}
\definecolor{paired-light-green}{RGB}{199, 233, 193}
\definecolor{paired-dark-green}{RGB}{49, 163, 83}
\definecolor{paired-light-purple}{RGB}{218, 218, 235}
\definecolor{paired-dark-purple}{RGB}{117, 107, 176}
\definecolor{paired-light-gray}{RGB}{217, 217, 217}
\definecolor{paired-dark-gray}{RGB}{99, 99, 99}
\definecolor{paired-light-pink}{RGB}{222, 158, 214}
\definecolor{paired-dark-pink}{RGB}{123, 65, 115}
\definecolor{paired-light-red}{RGB}{231, 150, 156}
\definecolor{paired-dark-red}{RGB}{131, 60, 56}
\definecolor{paired-light-yellow}{RGB}{231, 204, 149}
\definecolor{paired-dark-yellow}{RGB}{141, 109, 49}

\definecolor{forestgreen}{HTML}{2e7d43}
\definecolor{color1}{HTML}{FF9999}
\definecolor{color2}{HTML}{FF6666}
\definecolor{color3}{HTML}{FF3333}
\definecolor{color4}{HTML}{E60000}
\definecolor{color5}{HTML}{B30000}
\definecolor{color6}{HTML}{8CD98C}
\definecolor{color7}{HTML}{53c653}
\definecolor{color8}{HTML}{39ac39}
\definecolor{color9}{HTML}{2d862d}
\definecolor{color10}{HTML}{206020}
\definecolor{color11}{HTML}{cca300}

\definecolor{lightblue}{RGB}{212, 235, 255}
\definecolor{lightorange}{RGB}{255, 204, 168}
\definecolor{lightyellow}{RGB}{255, 255, 168}
\definecolor{lightgreen}{RGB}{224, 242, 213}
\definecolor{lightred}{RGB}{249,202,202}
\definecolor{lightgray}{RGB}{230,230,230}
\definecolor{critical}{HTML}{a90674}
\definecolor{major}{HTML}{f6736b}
\definecolor{minor}{HTML}{fae1af}
\definecolor{noerror}{HTML}{d2e8f1}

\usepackage[most]{tcolorbox}
\newtcbox{\hlmajortab}{on line, rounded corners, box align=base, colback=major!40,colframe=white,size=fbox,arc=3pt, before upper=\strut, top=-2pt, bottom=-4pt, left=-2pt, right=-2pt, boxrule=0pt}
\newtcbox{\hlcrittab}{on line, rounded corners, box align=base, colback=critical!40,colframe=white,size=fbox,arc=3pt, before upper=\strut, top=-2pt, bottom=-4pt, left=-2pt, right=-2pt, boxrule=0pt}
\newtcbox{\hlminortab}{on line, rounded corners, box align=base, colback=minor!40,colframe=white,size=fbox,arc=3pt, before upper=\strut, top=-2pt, bottom=-4pt, left=-2pt, right=-2pt, boxrule=0pt}
\newtcbox{\hlgraytab}{on line, rounded corners, box align=base, colback=gray!40,colframe=white,size=fbox,arc=3pt, before upper=\strut, top=-2pt, bottom=-4pt, left=-2pt, right=-2pt, boxrule=0pt}

\usepackage{graphicx}
\usepackage{caption}
\usepackage{subcaption}

\usepackage{arydshln}
\makeatletter
\def\adl@drawiv#1#2#3{%
        \hskip.5\tabcolsep
        \xleaders#3{#2.5\@tempdimb #1{1}#2.5\@tempdimb}%
                #2\z@ plus1fil minus1fil\relax
        \hskip.5\tabcolsep}
\newcommand{\cdashlinelr}[1]{%
  \noalign{\vskip 1.3pt
           \global\let\@dashdrawstore\adl@draw
           \global\let\adl@draw\adl@drawiv}
  \cdashline{#1}[.4pt/2pt]
  \noalign{\global\let\adl@draw\@dashdrawstore
           \vskip 1.3pt}}
\makeatother

\newcommand{\zh}[1]{\begin{CJK}{UTF8}{gbsn}#1\end{CJK}}

\title{\textbf{{\Large x}\textsc{comet}}: Transparent Machine Translation Evaluation through \\Fine-grained Error Detection}

\usepackage[misc]{ifsym}

\author{
    Nuno M. Guerreiro\thanks{~~Equal contribution. Corresponding authors: \\\Letter \, \url{{nuno.guerreiro, ricardo.rei}@unbabel.com}}\,\,$^{1,3,4,5}$,
    Ricardo Rei$^{*1,2,5}$,
    Daan van Stigt$^{1}$,
    \\
    \bf
    Luisa Coheur$^{2,5}$, Pierre Colombo$^{4}$, 
    André F. T. Martins$^{1,3,5}$
    \\
    $^{1}$Unbabel, Lisbon, Portugal, \,\ $^{2}$INESC-ID, Lisbon, Portugal \\
    $^{3}$Instituto de Telecomunicações, Lisbon, Portugal \\
    $^{4}$MICS, CentraleSupélec, Université Paris-Saclay, France\\
    $^{5}$Instituto Superior Técnico, University of Lisbon, Portugal\\[1.5ex]
}

\begin{document}
\maketitle
\begin{abstract}
Widely used learned metrics for machine translation evaluation, such as \textsc{Comet} and \textsc{Bleurt}, estimate the quality of a translation hypothesis by providing a single sentence-level score. As such, they offer little insight into translation errors (e.g., what are the errors and what is their severity). On the other hand, generative large language models (LLMs) are amplifying the adoption of more granular strategies to evaluation, attempting to detail and categorize translation errors. In this work, we introduce \textbf{x\textsc{comet}}, an open-source learned metric designed to bridge the gap between these approaches. \textbf{x\textsc{comet}} integrates both sentence-level evaluation and error span detection capabilities, exhibiting state-of-the-art performance across all types of evaluation (sentence-level, system-level, and error span detection). Moreover, it does so while highlighting and categorizing error spans, thus enriching the quality assessment. We also provide a robustness analysis with stress tests, and show that x\textsc{comet} is largely capable of identifying localized critical errors and hallucinations.

% However, achieving competitive performance to dedicated learned metrics through these strategies is still contingent on utilizing large \textit{closed} models, such as GPT-4 and PaLM-2.

% In this study, we introduce \textbf{x\textsc{comet}}\footnote{e\textbf{\textsc{x}}plainable \textbf{\textsc{C}}ross-lingual \textbf{\textsc{O}}ptimized \textbf{\textsc{M}}etric for \textbf{\textsc{E}}valuation of \textbf{\textsc{T}}ranslation}, a fresh neural fine-tuned metric that, in conjunction with a quality score, pinpoints fine-grained translation errors that closely align with the severity levels defined by MQM. Remarkably, \textbf{x\textsc{comet}} establishes a new state-of-the-art in terms of correlations with human judgments at the segment level, \textcolor{red}{no coma?} while presenting interpretable error spans that are in accordance with human annotations.

\end{abstract}

\section{Introduction}
Automatic metrics for machine translation evaluation are widely used by researchers and practitioners to evaluate the quality of translations and the systems generating them. Notably, \textit{learned} neural metrics, such as \textsc{Comet}~\cite{rei-etal-2020-comet} and \textsc{Bleurt}~\cite{sellam-etal-2020-bleurt}, have demonstrated significant improvements in terms of correlation with human judgements when compared to traditional metrics like \textsc{bleu}~\citep{papineni-etal-2002-bleu, freitag-etal-2021-results, freitag-etal-2022-results}.

These metrics are trained to regress on scores obtained through human annotations, by predicting a single sentence-level score representing the quality of the translation hypothesis. However, these single scores do not offer a detailed view into translation errors~(e.g., it is not immediate which words or spans of words are wrongly translated). Moreover, as they are obtained by making use of highly complex pre-trained models, they can be difficult to interpret~\citep{rei-etal-2023-inside, Leiter2023TowardsEE}. One appealing strategy to bring a more detailed view into translation errors is to obtain finer-grained information on error spans through highlighting them and indicating their severity~\citep{fonseca-etal-2019-findings, perrella-etal-2022-matese, bao2023finegrained}. In fact, this is the strategy adopted in recent works that have employed generative large language models~(LLMs) for machine translation evaluation: (i)~identify errors within a given translation, subsequently (ii)~categorize these errors according to their severity, and finally (iii)~infer a sentence-level score from the predicted errors~\cite{fernandes2023devil, xu2023instructscore}. However, these methods still lag behind dedicated learned metrics when using open LLMs, such as the LLaMA models~\cite{touvron2023llama, xu2023instructscore}. As it stands, competitive performance with generative strategies remains contingent on utilizing large \textit{proprietary}, \textit{closed} LLMs such as PaLM-2 and GPT-4~\cite{fernandes2023devil}.

\begin{figure*}
    \includegraphics[width=\textwidth]{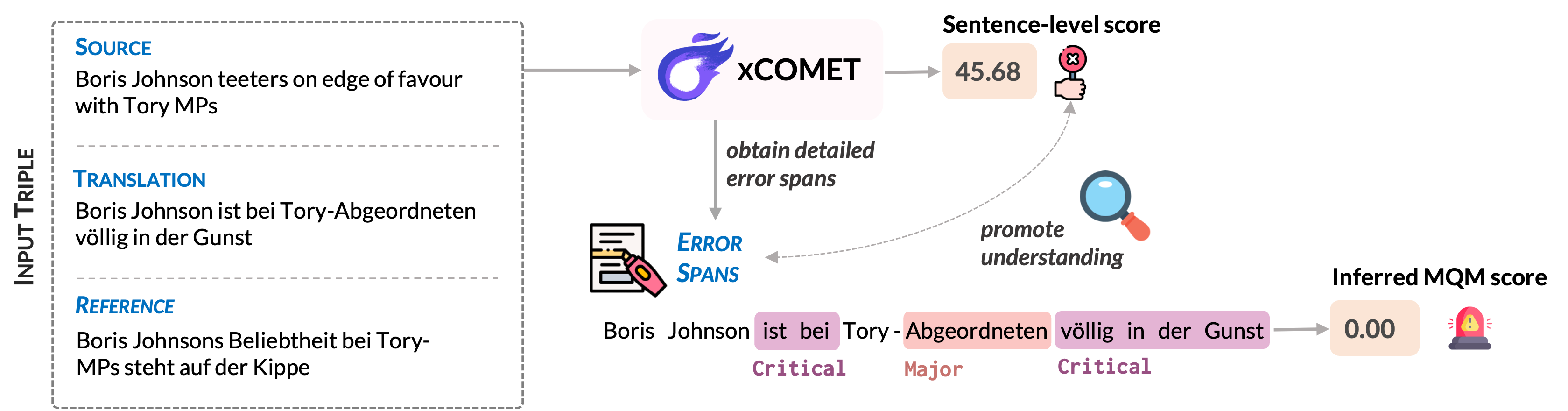}
    \caption{The \textbf{x\textsc{comet}} framework illustrated through a real example: the metric not only provides a sentence-level score, but also predicts translation error spans along with their respective severity. From these spans, we can infer MQM score (following the MQM typology) that informs and highly correlates with the sentence-level score~(see Section~\ref{sec:results}). These spans complement the sentence-level score by providing a detailed view into the translation errors.}
    \label{fig:xcomet_framework}
\end{figure*}

In this work, we bridge the gap between these two approaches to machine translation evaluation by introducing \textbf{x\textsc{comet}}: a \textit{learned} metric that simultaneously performs sentence-level evaluation and error span detection. Through extensive experiments, we show that our metrics leverage the strengths of both paradigms: they achieve state-of-the-art performance in all relevant vectors of evaluation (sentence-level, system-level, and error span prediction), while offering, via the predicted error spans, a lens through which we can analyze translation errors and better interpret the sentence-level scores. We achieve this by employing a curriculum during training that is focused on leveraging high-quality \textit{publicly available} data at both the sentence- and error span level, complemented by the construction of synthetic data to enhance the metric's robustness. Moreover, x\textsc{comet} is a unified metric~\citep{wan-etal-2022-unite}, accommodating all modes of evaluation within a single model. This enables the metric to be used even for quality estimation~(when no reference is available), or for reference-only evaluation, similarly to \textsc{Bleurt}~(when a source is not provided). Crucially, x\textsc{comet} also provides high-quality sentence-level scores that are directly inferred from the predicted error spans, in the style of \textsc{AutoMQM}~\cite{fernandes2023devil} and \textsc{InstructScore}~\cite{xu2023instructscore}.

Our contributions can be summarized as follows: 
\begin{enumerate}
    \item We introduce \textbf{x\textsc{comet}}, a novel evaluation metric that leverages the advantages of regression-based metrics and error span detection to offer a more detailed view of translation errors. 
    \item We show that \textbf{x\textsc{comet}} is a state-of-the-art metric at all relevant vectors of evaluation --- sentence-level, system-level, and error span prediction --- generally outperforming widely-used neural metrics and generative LLM-based machine translation evaluation.
    \item We provide a comprehensive robustness analysis of \textbf{x\textsc{comet}}, showing that this new suite of metrics identifies the vast majority of localized critical errors and hallucinations.
    \item We release two evaluation models: \textbf{x\textsc{comet-xl}}, with 3.5B parameters, and \textbf{x\textsc{comet-xxl}}, featuring 10.7B parameters.\footnote{The full suite of metrics (x\textsc{comet-xl} and x\textsc{comet-xxl}) will be released through HuggingFace Hub: \href{https://huggingface.co/Unbabel}{\texttt{https://huggingface.co/Unbabel}}.}
\end{enumerate}

\section{Background}
\paragraph{Methodologies for human assessment of translation quality.} Human evaluation of machine translation is primarily conducted through three distinct approaches: post-edits (PE), direct assessments~(DA), and the Multidimensional Quality Metrics~(MQM) framework. 

In post-edits (PE), \textit{professional translators} are tasked with ``fixing'' a given translation, making minimal edits to improve its quality. Using this edited translation --- often termed \textit{post-edit} --- we can evaluate the machine translation output by quantifying the number of edits, thus gauging the initial translation's quality~\citep{Snover06astudy}. 

Direct assessments (DA)~\cite{graham-etal-2013-continuous} are a simple and widely-used evaluation method. Annotators --- \textit{non-expert bilingual speakers} or \textit{professional translators} --- are asked to annotate each translation with a score ranging from 0 to 100 to reflect its adequacy and fluency, where a score of 100 corresponds to a perfect translation, and 0 corresponds to a completely inadequate one.

The Multidimensional Quality Metrics (MQM) framework \citep{lommel-mqm}, on the other hand, offers a more comprehensive and systematic approach to MT evaluation. \textit{Professional translators} highlight errors---typically in the form of error spans---~within translations, attributing them severity ratings (e.g., \textit{minor}, \textit{major}, or \textit{critical}) and categorical labels (e.g., \textit{fluency}, \textit{accuracy}). Figure~\ref{fig:xcomet_framework} illustrates one such annotation. MQM annotations have gained prominence in recent years due to their capacity to offer detailed insights into translation errors, facilitating more fine-grained and accurate comparisons between translation systems \citep{freitag-etal-2021-experts}. As such, the field of Automatic Evaluation of MT has increasingly favoured comparisons using MQM annotations over traditional DA and PE methodologies~\cite{freitag-etal-2021-results, freitag-etal-2022-results, zerva-etal-2022-findings}.

\paragraph{Automatic metrics for translation evaluation.}
Conventional automatic metrics for machine translation (MT) evaluation rely on \textit{lexical}-based approaches, where the evaluation score is computed through statistics related to lexical overlap between a machine translation and a reference translation. Despite evidence indicating that these lexical metrics (e.g., \textsc{Bleu}~\citep{papineni-etal-2002-bleu} and \textsc{chrF}~\citep{popovic-2015-chrf}) do not consistently align with human judgments, particularly when these are obtained through the MQM framework~\cite{freitag-etal-2021-results, freitag-etal-2022-results}, they remain very popular. In fact, \textsc{Bleu} remains the most widely employed evaluation metric in machine translation to this day~\cite{marie-etal-2021-scientific}. On the other hand, \textit{neural} metrics (e.g., \textsc{Comet}~\cite{rei-etal-2020-comet} and \textsc{Bleurt} \cite{sellam-etal-2020-bleurt}) that rely on complex neural networks to estimate the quality of MT outputs are consistently among the best metrics for MT evaluation according to correlations with human judgments~\cite{freitag-etal-2021-results, freitag-etal-2022-results}. 

However, contrary to lexical metrics which offer a straightforward interpretation, it can often prove challenging to explain the score predicted by a \textit{neural} metric to a given translation output. As such, there have been a series of efforts to bring interpretability to neural metrics by focusing on understanding the inner workings of neural metrics~\citep{rei-etal-2023-inside, Leiter2023TowardsEE}, or on constructing inherently interpretable neural metrics (e.g., \textsc{MaTESe} \citep{perrella-etal-2022-matese} and \textsc{Fg-ted} \cite{bao2023finegrained}) by assigning a central role to the task of predicting \textit{word-level} errors in a given translation, instead of \textit{just} a sentence-level score. 

More recently, with the rise of generative LLMs, some works have tried to frame the MT evaluation problem as a generative problem. This offers great flexibility, as the LLM can be prompted to either score the translation directly~\citep{kocmi-federmann-2023-large}, or to identify errors in the translation~(e.g., in line with the MQM framework)~\citep{fernandes2023devil, xu2023instructscore}.

\section{Problem Statement}
\label{sec:problem_statement}
An automatic \textit{metric} for translation evaluation aims at predicting the quality of a translated sentence, $\bm{t}$, in light of a reference translation, $\bm{r}$, for a given source sentence, $\bm{s}$. Here, we focus specifically on neural metrics that make use of a neural model, and typically operate under one of the following evaluation scenarios:
\begin{itemize}
\item \textbf{reference-only} (\textsc{ref}): the model evaluates the translation by processing it alongside a ground-truth reference sentence~(\textsc{Bleurt} is an example of such a metric);
\item \textbf{source-reference combined input} (\textsc{src+ref}): the model evaluates the translation by jointly processing it with both the source and the reference~(\textsc{Comet} is an example of such a metric);
\item \textbf{source-only} (\textsc{src}): the model evaluates the translation using only its corresponding source sequence~(\textsc{CometKiwi}~\citep{rei-etal-2022-cometkiwi} is an example of such a model). This mode is commonly termed as \textit{quality estimation}~(QE) or \textit{reference-free} evaluation~\citep{specia2010machine}.
\end{itemize}
In essence, the model's input sequence consists of the translation $\bm{t}$ paired with some \textbf{additional input}---either $\bm{r}$, $[\bm{r}, \bm{s}]$ or $\bm{s}$---derived from the scenarios above. Given this input, the model may predict the quality of the translation at different granularities, e.g., sentence-level or word(span)-level.

\paragraph{Sentence-level prediction.} The model is tasked to predict a single global score---typically between 0 and 1----for the translation, that represents how well it aligns with its context (i.e., source and/or reference sentence). These scores can be used for a broad range of tasks, such as gauging the quality of different translation systems~\citep{freitag-etal-2022-results}, identifying pathological translations~\citep{guerreiro-etal-2023-looking}, assisting the generation of translations by MT systems~\citep{fernandes-etal-2022-quality}, or even acting as reward models for human alignment of language models~\citep{gulcehre2023reinforced}. 

\paragraph{Word(span)-level prediction.} In contrast, word-level~(or span-level) predictions are more fine-grained, identifying individual words or phrases in the translation that may have errors or discrepancies---typically identifying them as \textsc{ok}/\textsc{bad} or according to their severity, e.g., \textsc{minor}/\textsc{major}. These granular evaluations are more interpretable and assist in pinpointing specific issues, which can be particularly valuable for feedback and iterative translation improvements.

Our metric, \textbf{x\textsc{comet}}, emerges in a unique position in the landscape of MT evaluation metrics. It can simultaneously perform evaluation under all of the three scenarios (\textsc{src}, \textsc{ref}, \textsc{src+ref}) presented, and provide sentence-level scores and error span annotations that are in line with the MQM framework, thus bringing further transparency to the evaluation~(see Figure~\ref{fig:xcomet_framework} for an illustration). In the next section, we detail the design choices and methodology of x\textsc{comet}.

\section{Design and Methodology of x\textsc{comet}}
In this section, we describe the methodology behind x\textsc{comet}, outlining its model architecture, training settings and corpora, and learning curriculum. We detail how the model is designed to perform both regression and error span detection while adopting a unified input approach for enhanced flexibility and performance. 

\subsection{Model Architecture}
\label{ssec:architecture}

\definecolor{paired-light-blue}{RGB}{198, 219, 239}
\definecolor{paired-dark-blue}{RGB}{49, 130, 188}
\definecolor{paired-light-orange}{RGB}{251, 208, 162}
\definecolor{paired-dark-orange}{RGB}{230, 85, 12}
\definecolor{paired-light-green}{RGB}{199, 233, 193}
\definecolor{paired-dark-green}{RGB}{49, 163, 83}
\definecolor{paired-light-purple}{RGB}{218, 218, 235}
\definecolor{paired-dark-purple}{RGB}{117, 107, 176}
\definecolor{paired-light-gray}{RGB}{217, 217, 217}
\definecolor{paired-dark-gray}{RGB}{99, 99, 99}
\definecolor{paired-light-pink}{RGB}{222, 158, 214}
\definecolor{paired-dark-pink}{RGB}{123, 65, 115}
\definecolor{paired-light-red}{RGB}{231, 150, 156}
\definecolor{paired-dark-red}{RGB}{131, 60, 56}
\definecolor{paired-light-yellow}{RGB}{231, 204, 149}
\definecolor{paired-dark-yellow}{RGB}{141, 109, 49}
\tikzset{%
    layernode/.style = {
        align=center,
        text width=6cm,
        inner sep=0.25cm,
        outer sep=0cm,
        rounded corners=4pt,
        fill=paired-light-gray!30,
        draw=paired-dark-gray!50,
    },
    halfnode/.style = {
        align=center,
        text width=2.5cm,
        inner sep=0.25cm,
        outer sep=0cm,
        rounded corners=4pt,
        fill=paired-light-gray!30,
        draw=paired-dark-gray!50,
    },
    outputnode/.style = {
        align=center,
        text width=2.6cm,
        % inner sep=0.25cm,
        % outer sep=0cm,
        % rounded corners=4pt,
        % fill=paired-light-orange!30,
        % draw=paired-dark-orange!45,
    },
    inputnode/.style = {
        align=center,
        % rounded corners=3pt,
        inner sep=0.2cm,
    },
}

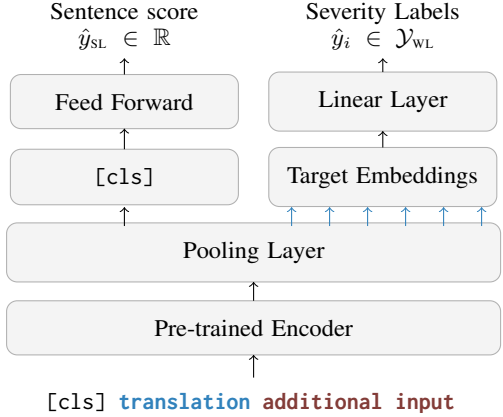
\begin{figure}[t]
    \centering
    \small
    \begin{tikzpicture}
        %Nodes
        \node[inputnode] (input) at (0, 0) {
            \texttt{
            [cls] \textcolor{paired-dark-blue}{\textbf{translation}}  \textcolor{paired-dark-red}{ \textbf{additional input}}
        }
        };
        \node[layernode] (encoder) at (0, 1) {Pre-trained Encoder};
        \node[layernode] (scalarmix) at (0, 2) {Pooling Layer};
        % \node[halfnode] (avgpool) at (-1.7, 3) {Avg. Pooling};
        \node[halfnode] (avgpool) at (-1.7, 3) {\texttt{[cls]}};
        \node[halfnode] (piece) at (1.7, 3) {Target Embeddings};
        \node[halfnode] (ffn1) at (-1.7, 4) {Feed Forward};
        \node[halfnode] (ffn2) at (1.7, 4) {Linear Layer};
        \node[outputnode] (out1) at (-1.7, 5) {Sentence score \\ $\hat{y}_{\textsc{sl}} \in \mathbb{R}$};
        \node[outputnode] (out2) at (1.7, 5) {Severity Labels \\ $\hat{y}_i \in \mathcal{Y}_{\textsc{wl}}$};
        
        %Lines
        \draw[->] (input) to (encoder);
        \draw[->] (encoder) to (scalarmix);
        % \draw[->] (scalarmix) to (avgpool);
        % \draw[->,paired-dark-red] (-0.5, 2.35) to (-0.5, 2.6);
        % \draw[->,paired-dark-red] (-1.0, 2.35) to (-1.0, 2.6);
        % \draw[->,paired-dark-red] (-1.5, 2.35) to (-1.5, 2.6);
        \draw[->] (-1.7, 2.35) to (-1.7, 2.6);
        % \draw[->,paired-dark-blue] (-2.0, 2.35) to (-2.0, 2.6);
        % \draw[->,paired-dark-blue] (-2.5, 2.35) to (-2.5, 2.6);
        % \draw[->,paired-dark-blue] (-3.0, 2.35) to (-3.0, 2.6);
        % \draw[->] (scalarmix) to (piece);
        \draw[->,paired-dark-blue] (0.5, 2.35) to (0.5, 2.6);
        \draw[->,paired-dark-blue] (1.0, 2.35) to (1.0, 2.6);
        \draw[->,paired-dark-blue] (1.5, 2.35) to (1.5, 2.6);
        \draw[->,paired-dark-blue] (2.0, 2.35) to (2.0, 2.6);
        \draw[->,paired-dark-blue] (2.5, 2.35) to (2.5, 2.6);
        \draw[->,paired-dark-blue] (3.0, 2.35) to (3.0, 2.6);
        \draw[->] (avgpool) to (ffn1);
        \draw[->] (piece) to (ffn2);
        \draw[->] (ffn1) to (out1);
        \draw[->] (ffn2) to (out2);

    \end{tikzpicture}
    
    \caption{
    Architecture of \textbf{x\textsc{comet}}. The input to the model starts with a \texttt{[cls]} token followed by a \textcolor{paired-dark-blue}{\texttt{\textbf{translation}}} and an \textcolor{paired-dark-red}{\texttt{\textbf{additional input}}} that will have the source, reference or both. After the pooling layer the \texttt{[cls]} token is passed to a feed-forward to produce a quality score while all subword pieces corresponding to the \textcolor{paired-dark-blue}{\texttt{\textbf{translation}}} are passed to a linear layer that will classify them according to their severity levels, $\mathcal{Y}_{\textsc{wl}} = \{\textsc{ok}, \textsc{min}, \textsc{maj},\textsc{crit}\}$.
    }
    \label{fig:arch}
\end{figure}

x\textsc{comet} is built upon insights garnered from Unbabel-IST's contributions to the WMT22 Metrics and QE shared tasks~\cite{rei-etal-2022-comet, rei-etal-2022-cometkiwi}. It is designed to concurrently handle two tasks:~sentence-level regression and error span detection. Figure \ref{fig:arch} illustrates its architecture. We follow the same architecture of the scaled-up version of \textsc{CometKiwi} detailed in~\citet{rei2023scaling}, which uses a large pre-trained encoder model as its backbone encoder model. Importantly, following naturally from our multi-task setup, the model has two prediction heads: (i)~a sentence-level \textit{regression} head, which employs a feed-forward network to generate a sentence score, and (ii)~a word-level \textit{sequence tagger}, which applies a linear layer to assign labels to each token in a given translation.

We train two \textbf{x\textsc{comet}} versions --- \textbf{x\textsc{comet-xl}} and \textbf{x\textsc{comet-xxl}} --- using the XL (3.5B parameters) and XXL (10.7B parameters) versions of XLM-R~\cite{goyal-etal-2021-larger}.\footnote{To the best of our knowledge, these represent the two largest open-source encoder-only models.}

\subsection{Fully Unified Evaluation}
x\textsc{comet} adopts a unified input approach \cite{wan-etal-2022-unite}, allowing for all the evaluation scenarios described in Section~\ref{sec:problem_statement}---\textsc{ref}, \textsc{src+ref}, and \textsc{src} evaluation---under a single model. Thus, the input sequence consists of two parts: (i)~the translated sentence $\bm{t} = [t_1, ..., t_n]$ of length $n$, and (ii)~an additional input containing information from the source, reference, or both.\footnote{Each input is always delimited by separators. For example, the unified input can be written as: \textcolor{paired-dark-blue}{\texttt{<s>}\textbf{\texttt{translation}}\texttt{<s>}} \textcolor{paired-dark-red}{\texttt{</s>\textbf{src}</s></s>\textbf{ref}</s>}}} To do so, when a reference is available, we run three distinct forward passes~(one for each evaluation scenario), each yielding sentence-level and word-level predictions.

\subsubsection{Training time}
For each forward-pass, we collect the sentence-level predictions $\{\hat{y}_{\textsc{sl}}^{\textsc{src}}, \hat{y}_{\textsc{sl}}^{\textsc{ref}}, \hat{y}_{\textsc{sl}}^{\textsc{src+ref}}\}$ and the word-level logits $\{\bm{\hat{y}}_{\textsc{wl}}^{\textsc{src}}, \bm{\hat{y}}_{\textsc{wl}}^{\textsc{ref}}, \bm{\hat{y}}_{\textsc{wl}}^{\textsc{src+ref}}\}$.\footnote{Here, for each \textsc{input}\,$\in$\,\{\textsc{src}, \textsc{ref}, \textsc{src+ref}\}, we define $\bm{\hat{y}}_{\textsc{wl}}^{\textsc{Input}} = [\hat{y}_1^{\textsc{Input}}, \dots, \hat{y}_n^{\textsc{Input}}]$.}

As we have mentioned before, x\textsc{comet} models are trained with supervision from both sentence-level quality assessments, $y_{\textsc{sl}}$, and word-level severity tags, $\bm{y}_{\textsc{wl}} = [{y}_1, \dots, {y}_n]$, with ${y}_i \in \mathcal{Y}_{\textsc{wl}} = \{\textsc{ok}, \textsc{min}, \textsc{maj},\textsc{crit}\}$. In the multi-task setting, we use the following loss $\mathcal{L}$ for each input type~(\textsc{input}\,$\in$\,\{\textsc{src}, \textsc{ref}, \textsc{src+ref}\}):
\begin{align}
    \mathcal{L}_{\textsc{sl}}^{\textsc{input}} &=  \left( y_{\textsc{sl}} - \hat{y}_{\textsc{sl}}^{\textsc{input}} \right)^2 \\
    \mathcal{L}_{\textsc{wl}}^{\textsc{input}} &= -\frac{1}{n}\sum_{i=1}^{n} \alpha_{y_i} \log p(\hat{y}_i^{\textsc{input}}) \\
    \mathcal{L}^{\textsc{input}} &= (1-\lambda) \mathcal{L}_{\textsc{sl}}^{\textsc{input}}+ \lambda \mathcal{L}_{\textsc{wl}}^{\textsc{input}}, \label{eq:comb_loss}
\end{align}
$\bm{\alpha} \in \mathbb{R}^{|\mathcal{Y}_{\textsc{wl}}|}$ represents the class weights given for each severity label and $\lambda$ is used to weigh the combination of the sentence and word-level losses.

The final learning objective is the summation of the losses for each input type:
\begin{align}
    \mathcal{L} = \mathcal{L}^{\textsc{src}} + \mathcal{L}^{\textsc{ref}} + \mathcal{L}^{\textsc{src+ref}}
\end{align}

Furthermore, in line with preceding metrics constructed upon the \textsc{Comet} framework, our models use features such as gradual unfreezing, and discriminative learning rates. See Appendix~\ref{app_sec:hyperparameters} for full details and hyperparameters.

\subsubsection{Inference time}

\paragraph{Error span prediction.} For each subword in the translation, we average the output distribution of the word-level linear layer obtained for each forward pass. Using this distribution, we predict a set of word-level tags $\bm{\hat{y}}_{\textsc{wl}} = [\hat{y}_1, \dots, \hat{y}_n]$. From these tags, we construct a list of \textit{error spans}, $S$, by grouping adjacent subwords identified as errors. The severity of each span in $S$ is defined according to the most severe error tag found within the span.

\paragraph{Sentence-level prediction.} For each forward pass, we obtain the corresponding sentence-level scores: $\hat{y}_{\textsc{src}}$, $\hat{y}_{\textsc{ref}}$, and $\hat{y}_{\textsc{src+ref}}$. Additionally, we leverage the information coming from the predicted list of error spans, $S$, to infer an automated MQM score. To do so, we follow the MQM framework: we obtain the error counts for each severity level---~$c_{\textsc{min}}, c_{\textsc{maj}}, c_{\textsc{crit}}$~---and apply the predetermined severity penalty multipliers to define the error type penalty total, $e(S)$. Formally:
\begin{equation}\label{eq:error_count}
    e(S) = c_{\textsc{min}} + 5\times c_{\textsc{maj}} + 10\times c_{\textsc{crit}}.
\end{equation}
Finally, we obtain $\hat{y}_{\textsc{mqm}}$ by capping and flipping the sign of $e(S)$:
\begin{equation}\label{eq:automqm}
  \hat{y}_{\textsc{mqm}} =\begin{cases}
    \frac{25 - e(S)}{25}, & \text{if $e(S) < 25$}.\\
    0, & \text{otherwise}.
  \end{cases}
\end{equation}
Note that the predicted score $\hat{y}_{\textsc{mqm}}$ is bounded between 0 and 1, with a score of 1 corresponding to a perfect translation.

We aggregate the scores to compute the final sentence-level score, $\hat{y}_{\textsc{sl}}$, through a weighted sum of the different sentence-level scores. Importantly, we also include the inferred MQM score $\hat{y}_{\textsc{mqm}}$ to directly inform the final sentence-level prediction. Formally, given $\bm{\hat{y}} = [\hat{y}_{\textsc{src}}, \hat{y}_{\textsc{ref}}, \hat{y}_{\textsc{src+ref}}, \hat{y}_{\textsc{mqm}}]$:
\begin{equation}
\label{eq:y_sl_computation}
\hat{y}_{\textsc{sl}} = \bm{w}^\top \bm{\hat{y}}
\end{equation}
where $\bm{w}$ is set to $[1/9, 1/3, 1/3, 2/9]$.\footnote{\textsc{UniTE} uniformly distributes the weight across the different sentence-level scores to obtain the final prediction. However, we found that, in practice, distributing the weight of each sentence-level prediction can lead to improved results.}

\subsection{Corpora}
\label{ssec:corpora}

Our models are exclusively trained on publicly available DA and MQM annotations, most of which have been collected by WMT over the recent years. 

\paragraph{DA data.} We use DA annotations collected by WMT from 2017 to 2020, and the MLQE-PE dataset \citep{fomicheva-etal-2022-mlqe}. As the MLQE-PE dataset does not contain reference translations, we used the post-edit translations as reference translations. Overall, the corpus consists of around 1 million samples, spanning 36 language pairs.

\paragraph{MQM data.} We collected the MQM annotations sourced from WMT from 2020 to 2022.\footnote{Here, we exclude the 2022 News domain annotations, which we reserved for testing.} We also used annotations sourced from other MQM-annotated datasets: (i)~IndicMT~\cite{sai-b-etal-2023-indicmt}, which contains MQM annotations spanning 5 Indian languages, and (ii)~DEMETR~\cite{karpinska-etal-2022-demetr}, a diagnostic dataset with perturbations spanning semantic, syntactic, and morphological error categories.

Corpora with MQM annotations are usually extremely unbalanced with critical errors being underrepresented. In term, this may lead to metrics dealing less well with pathological translations, such as critical errors and hallucinations~\citep{amrhein-sennrich-2022-identifying, raunak-etal-2022-salted, guerreiro-etal-2023-looking}. As such, we augment the MQM corpus with \textit{synthetic critical} errors. We create different types of detached and oscillatory hallucinations~\citep{raunak-etal-2021-curious, guerreiro-etal-2023-looking}: (i) detached hallucinations, where we replace the translation with a random sentence; (ii) other detached hallucinations, where we replace the true translation with an unrelated translation that is semantically similar to the source sentence\footnote{We measure cross-lingual similarity via the cosine similarity between the sentence embeddings obtained with the LaBSE encoder~\citep{feng-etal-2022-language}.}; and (iii)~oscillatory hallucinations, where we randomly sample a $n$-gram from the translation~(with $n$ in $\{2,3,4\}$) and repeat it between 1 and 10 times. We provide examples of these synthetic hallucinations in Appendix~\ref{app_sec:synthetic_examples_halls}. Overall, our MQM corpus consists of 176K samples, spanning 14 language pairs.

% 10\% ``critical'' translations from three categories: random translations, where we replace the MT with a random sentence; similar source translations, where we replace the MT with a translation from a similar source sentence; and repetitions, where we randomly repeat n-grams in the translation. For the first two types, we consider the entire translation as a single critical error and set the score to 0. For repetitions, the inserted text is marked as a critical error, and we compute the score following Eq. \ref{eq:automqm}.
 
% Each year since 2017, the organizers of the WMT News Translation Shared Task \cite{barrault-etal-2019-findings} have gathered human judgments in the form of DA \citep{graham-etal-2013-continuous}. These judgments are subsequently converted into z-scores to mitigate variations in scoring strategies among different annotators \cite{bojar-etal-2017-findings}. Notably, the quality annotations adopted by the WMT QE shared task led to the creation of the MLQE-PE dataset \citep{fomicheva-etal-2022-mlqe}. As a result, our DA corpus comprises DA annotations from 2017, 2018, 2019, 2020, and the MLQE-PE dataset \citep{fomicheva-etal-2022-mlqe}. Although the MLQE-PE corpus lacked explicit references, we utilized the post-edit translations as reference translations. The resulting corpus encompasses 1,027,155 tuples, spanning 36 language pairs (language distribution is outlined in Appendix X\todo{missing}).

\paragraph{Scaling of sentence-level scores.} While the sentence-level scores inferred from MQM annotations~(through the procedure in Equation~\ref{eq:automqm}) are bounded between 0 and 1, DA annotations usually require $z$-normalization in order to mitigate variations in scoring strategies by different annotators~\cite{bojar-etal-2017-findings}.\footnote{This is particularly relevant for DA annotations, since these judgements typically come from non-expert annotators.} Thus, as $z$-scores are inherently centered at 0 and unbounded, there is a scaling mismatch between the data samples. 

Consequently, to circumvent this limitation, we employ min-max scaling on our DA corpus to set its range of scores to $[0,1]$. To do so, we set a practical minimum and maximum $z$-score value. We obtain the minimum score by averaging the $z$-scores for translations with over 1 annotation, wherein all annotators unanimously scored them with an unnormalized 0 DA score, i.e., they deemed the translation as ``random''. For determining a maximum value, we applied the same process for perfect translations, i.e., unnormalized 100 DA score.\footnote{This technique was initially introduced in \textsc{Bleurt-20} \cite{pu-etal-2021-learning}.}

% \subsubsection{MQM annotations}
% In contrast to the readily available abundance of DA data in the literature, publicly accessible MQM annotations remain relatively constrained in availability. Mirroring our approach with DA data, we sought to consolidate diverse sources of publicly available annotations into a unified MQM corpus.
% \paragraph{Data sources.} Our MQM corpus encompasses annotations derived from the WMT 2021 and 2022 shared tasks (with the exception of the 2022 News domain annotations, reserved for testing), annotations referenced in \citep{freitag-etal-2021-experts}, and annotations sourced from IndicMT \cite{sai-b-etal-2023-indicmt}. Moreover, we enhance this corpus through the incorporation of perturbations from Demetr, aligned with an MQM typology \cite{karpinska-etal-2022-demetr}. This concerted effort results in a comprehensive MQM corpus comprising 176K samples, spanning 14 language pairs. 
% The resulting sentence-level MQM score, following Eq. \ref{eq:automqm}, adheres to a range between 0 and 1, thus facilitating a seamless transfer of information between the DA and MQM corpora.

\subsection{Training Curriculum}
\label{ssec:training}

\textbf{x\textsc{comet}} models undergo a 3-phase curriculum training. Throughout these phases, the training emphasis alternates between sentence-level prediction and error span prediction by tweaking the parameter $\lambda$ in Equation~\ref{eq:comb_loss}. The curriculum phases can be described as follows: 

\begin{description}[leftmargin=0em]
    \item[Phase I:] The model is trained exclusively using the DA data. In this phase, the focus is exclusively set on sentence-level regression.
    \item[Phase II:]  In this stage, we introduce word-level supervision. To achieve this, the model is fine-tuned on our diverse MQM corpus, with most emphasis placed on the word-level task.
    \item[Phase III:] The last training phase is aimed at unifying both tasks. The model is further fine-tuned using high-quality MQM data from \cite{freitag-etal-2021-experts}, with a bigger emphasis set to sentence-level prediction.
\end{description}

We describe how we obtain the values of $\lambda$ for Phases II and III in Appendix~\ref{app_sec:hyperparameters}.\footnote{The achieved $\lambda$ weights for Phases II and III were $\lambda = 0.983$ and $\lambda = 0.055$, respetively.}

\paragraph{Interpretation of the curriculum.} We first start by training a sentence-level metric --- similar to \textsc{UniTE}~\citep{wan-etal-2022-unified} --- on the vastly available DA annotations. This first phase acts as a warm-up for subsequent stages. In fact, prior research has shown that models trained on DA annotations leverage token-level information that aligns with MQM error annotations~\cite{rei-etal-2023-inside}. When we move to the second phase, we assume that we have a metric that can perform sentence-level regression. Thus, the aim here shifts to integrating word-level supervision without compromising the previously acquired sentence-level prediction skills. To do so, we use the highly diverse corpora of MQM annotations and set most emphasis on the word-level task. Finally, we exclusively leverage a small corpus~(around 25k samples) of very high-quality MQM annotations from~\cite{freitag-etal-2021-experts}  ---~each sample has three annotations from separate annotators ---~with additional synthetic hallucinations. Our focus here is to mitigate any potential decline in sentence-level regression capabilities during Phase II. %The values of $\lambda$ for each phase and the procedure to obtain them is \textcolor{red}{are} described in Appendix~\ref{app_sec:hyperparameters}.

\section{Experimental Setting}

\begin{table*}[ht!]
\centering
\footnotesize
\begin{tabular}{lrrrrrrrrrrr}
\toprule
   & \multicolumn{2}{c}{\texttt{zh-en}} & & \multicolumn{2}{c}{\texttt{en-de}}  & &\multicolumn{2}{c}{\texttt{en-ru}} & &\multicolumn{2}{c}{\texttt{Avg.}}\\\cmidrule{2-3}\cmidrule{5-6}\cmidrule{8-9}\cmidrule{11-12}
   \textsc{Metric}  & \multicolumn{1}{c}{$\rho$} & \multicolumn{1}{c}{$\tau$}  & & \multicolumn{1}{c}{$\rho$}  & \multicolumn{1}{c}{$\tau$} &  & \multicolumn{1}{c}{$\rho$} & \multicolumn{1}{c}{$\tau$} & & \multicolumn{1}{c}{$\rho$} & \multicolumn{1}{c}{ $\tau$} \\ \midrule
   % \multicolumn{12}{c}{{\textit{Baselines}}} \\
\multicolumn{1}{l}{\textsc{Bleurt-20}} & 0.462 & 0.336 & & 0.568 & 0.380 & & 0.498 & 0.379 & & 0.509 & 0.365 \\ 
\multicolumn{1}{l}{\textsc{comet-22}} & 0.423 & 0.335 & & 0.581 & 0.369 &  & 0.516 & 0.391  & & 0.507 & 0.361 \\ 
\multicolumn{1}{l}{\textsc{MetricX}} & \textbf{0.573} & \textbf{0.415} & &\textbf{0.640} & 0.405 & & 0.581	& 0.444 & &0.598 & 0.421  \\
\multicolumn{1}{l}{\textsc{Gemba-gpt4-da}$^\star$} & 0.318 & 0.292 & & 0.508 & 0.387 & & 0.454	& 0.383 & & 0.427 & 0.354  \\\midrule
% \multicolumn{12}{c}{\textbf{\textit{Ours}}} \\
\multicolumn{1}{l}{x\textsc{comet-xl}} & 0.556 & 0.399 & &\textbf{0.653} & 0.414 & & 0.611 & 0.448 & & 0.607 & 0.420 \\
\multicolumn{1}{l}{x\textsc{comet-xxl}} & 0.554	& 0.390 & & \textbf{0.644} & \textbf{0.435} & & \textbf{0.628}	& \textbf{0.470} & & \textbf{0.609} &	\textbf{0.432}\\\cdashlinelr{1-12}
\multicolumn{12}{c}{\textit{Predicted MQM scores from the error spans ($\hat{y} = \hat{y}_{\textsc{mqm}}$)}} \\
\multicolumn{1}{l}{{x\textsc{comet-xl}} (\textsc{mqm})} & 0.447 & 0.374 & & 0.561 & 0.389 & & 0.534 & 0.445 & & 0.514 &	0.402 \\
\multicolumn{1}{l}{{x\textsc{comet-xxl}} (\textsc{mqm})} & 0.446 & 0.332 & & 0.597 & 0.415 & & 0.533 & 0.439 & & 0.525 & 0.395 \\\hline
\end{tabular}
\caption{Segment-level Pearson ($\rho$) and Kendall-Tau ($\tau$)~($\uparrow$) using the Perm-Both hypothesis test~\citep{deutsch-etal-2021-statistical}. Numbers in bold belong to the top-performing cluster according to statistical significance~($p<0.05$).}
\label{tab:seg-results}
\end{table*}    

\subsection{Evaluation}

We test our metrics on the MQM annotations from the News domain from the WMT 2022 Metrics shared task. These annotations cover three language pairs: Chinese$\to$English (\texttt{zh-en}), English$\to$German (\texttt{en-de}), and English$\to$Russian (\texttt{en-ru}).\footnote{The test set comprises 4,500 segments for \texttt{en-de}, 4,500 for \texttt{en-ru}, and 7,575 for \texttt{zh-en}, sourced from 15 different translation systems.} We evaluate the metrics in terms of sentence-level, system-level, and error span prediction performance.

At the sentence-level, we report both the Pearson correlation coefficient ($\rho$) and Kendall's Tau ($\tau$) using the Perm-Both hypothesis test \citep{deutsch-etal-2021-statistical}. We also evaluate the metrics on System-level Pairwise Accuracy~\citep{kocmi-etal-2021-ship}. We base these evaluations on 200 re-sampling runs, with a significance level~($p$) set to 0.05. For error span prediction, we adopt the WMT23 Quality Estimation shared task evaluation methodology and compute F1 scores calculated at the character level, taking into account partial matches for both minor and major errors.\footnote{We convert all critical errors into major errors, in order to match the guidelines~(focused exclusively on minor and major errors) described in \cite{freitag-etal-2021-experts}, that were used for annotating the \texttt{zh-en} and \texttt{de-en} test sets.}

\subsection{Baselines}

\paragraph{Sentence and system-level.} We benchmark our metrics widely used \textit{open} neural metrics: \textsc{Comet-22}~\citep{rei-etal-2022-comet}\footnote{Used checkpoint: \texttt{Unbabel/wmt22-comet-da}} and \textsc{Bleurt-20}~\citep{pu-etal-2021-learning}. Additionally, we include \textsc{MetricX}, the best performing metric from WMT22 Metrics shared task~\citep{freitag-etal-2022-results}.\footnote{Specifically, we employ the \texttt{metricx\_xxl\_MQM\_2020} submission scores from the \texttt{mt-metrics-eval} package. Although the metric has not been released publicly, it is public that it is built upon the mT5-XXL \cite{xue-etal-2021-mt5} and has 13B parameters~\citep{deutsch2023training}.} Finally, we also include \textsc{Gemba} \cite{kocmi-federmann-2023-large}, which employs \textsc{GPT4}~\citep{openai2023gpt4} to evaluate translations following DA guidelines.

\paragraph{Error span prediction.} We report results using GPT3.5 and GPT4 models, by prompting it in the style of \textsc{AutoMQM}~\cite{fernandes2023devil}.\footnote{We use the models from the OpenAI API~(\texttt{gpt-3.5-turbo} and \texttt{gpt-4}) in October, 2023.} We carefully select 5 shots that are held constant for all samples. This way, we can directly compare our results with state-of-the-art LLMs, which have been shown to be able to perform the task of error detection~\cite{fernandes2023devil, xu2023instructscore}.

\section{Correlations with Human Judgements}
\label{sec:results}

\begin{table}[t]
\centering
\footnotesize
\setlength{\tabcolsep}{.4em}
\begin{tabular}{lcccc}
\toprule
\textsc{Metric} & \texttt{zh-en} & \texttt{en-de} & \texttt{en-ru} & \texttt{Avg.} \\ \midrule
\textsc{Bleurt-20} & 0.762 & 0.771 & 0.743 & 0.759 \\
\textsc{comet-22} & 0.705 & 0.800 & 0.733 & 0.746 \\
\textsc{MetricX} & 0.762 & 0.781 & 0.724 & 0.756 \\ 		
\textsc{Gemba-gpt4-da} & 0.752 & \textbf{0.848} & \textbf{0.876} & \textbf{0.825} \\\midrule
x\textsc{comet-xl} & \textbf{0.800} & 0.743 & 0.790 & 0.778 \\
x\textsc{comet-xxl} & \textbf{0.800} & \textbf{0.829} & \textbf{0.829} & \textbf{0.819}\\\cdashlinelr{1-5}
\multicolumn{5}{c}{\textit{MQM scores from the error spans ($\hat{y} = \hat{y}_{\textsc{mqm}}$)}} \\
x\textsc{comet-xl} (\textsc{mqm}) & 0.781 & 0.762 & 0.762 & 0.768 \\
x\textsc{comet-xxl} (\textsc{mqm}) & 0.781 & \textbf{0.838} & 0.810 & 0.810\\\bottomrule
\end{tabular}
\caption{System-level Pairwise Accuracy ($\uparrow$)~\citep{kocmi-etal-2021-ship} using the Perm-Both hypothesis test \citep{deutsch-etal-2021-statistical}. Numbers in bold belong to the top-performing cluster according to statistical significance~($p<0.05$).}
\label{tab:sys-results} 
\end{table}

In this section, we present a standard performance analysis of our metrics in terms of correlations with human judgments. Overall, we find x\textsc{comet} to be a state-of-the-art in sentence-level and error span prediction, being competitive with generative LLMs in terms of system-level evaluation.

\paragraph{Sentence-level evaluation.} Table \ref{tab:seg-results} shows that both x\textsc{comet} metrics outperform other strong performing neural metrics, including the generative approach leveraging GPT4 of \textsc{Gemba}. In particular, x\textsc{comet-xxl} sets a new state-of-the-art for \texttt{en-de} and \texttt{en-ru}. Interestingly, we can see that, while scaling up the encoder model of the x\textsc{comet} metrics~(from \textsc{xl} to \textsc{xxl}) holds better results, x\textsc{comet-xl} is very competitive.\footnote{Note that this corresponds to an increase in the parameter size of over 7 billion parameters.} In fact, it outperforms \textsc{MetricX}, which runs at even a larger size than x\textsc{comet-xxl}. Finally, we can also observe that the MQM scores inferred exclusively from the predicted error spans also exhibit strong performance, outperforming widely used metrics \textsc{Bleurt-20} and \textsc{Comet-22}. This is particularly relevant: the predicted error spans bring not only a more detailed view into translation errors but also provide high-quality sentence-level scores.

\paragraph{System-level evaluation.} Table~\ref{tab:sys-results} shows results for system-level. Similarly to what we observed at the sentence-level, our metrics show consistently superior performance when compared to other dedicated neural metrics. Notably, although generative approaches typically do much better at system-level evaluation when compared to dedicated models~\citep{kocmi-federmann-2023-large, fernandes2023devil}, x\textsc{comet-xxl} remains competitive in all language pairs with \textsc{Gemba} using GPT4. Finally, building on the findings at the sentence-level, the MQM scores inferred directly and exclusively from the predicted error spans also exhibit very competitive performance in terms of system-level accuracy. 

\paragraph{Error span prediction.} While we have highlighted the utility of the predicted error spans through the inferred sentence-level MQM scores, here we turn to evaluating them directly. Table \ref{tab:results_word_level} shows that the error spans predicted via x\textsc{comet} metrics outperform those obtained with both GPT3.5 and GPT4 despite being smaller in capacity relative to these models. In fact, our metrics achieve close performance to that of GPT4, even when a reference is not provided. 

\begin{table}[t]
    \centering
    \small    
    \begin{tabular}{l ccccc}
        \toprule
        \textsc{Metric} & \texttt{zh-en} & \texttt{en-de} & \texttt{en-ru} & \texttt{Avg.}  \\
        \midrule
        \textcolor{frenchblue!50}{$\bullet$} AutoMQM (\textsc{gpt}3.5) & 0.143 & 0.160 & 0.166 & 0.156 \\
        \textcolor{frenchblue!50}{$\bullet$} AutoMQM (\textsc{gpt}4) & 0.248 & 0.257 & \textbf{0.281} & 0.262 \\
        \midrule
        \textcolor{frenchblue!50}{$\bullet$} {x\textsc{comet-xl}} & 0.237 & 0.290 & \textbf{0.281} & 0.269 \\
        \textcolor{frenchblue!50}{$\bullet$} {x\textsc{comet-xxl}} & \textbf{0.257} & \textbf{0.320} & 0.262 & \textbf{0.280} \\\cdashlinelr{1-6}
        \multicolumn{5}{c}{\textit{Error spans detected with source-only}}\\
        \textcolor{camel!60}{$\bullet$} {x\textsc{comet-xl}} (\textsc{src}) & 0.208 & 0.264 & 0.252 & 0.242 \\ \textcolor{camel!60}{$\bullet$} {x\textsc{comet-xxl}} (\textsc{src}) & 0.229 & 0.298 & 0.238 & 0.255 \\ 
        \bottomrule
    \end{tabular}
    \caption{F1 scores ($\uparrow$) on error span detection for reference-free (\textcolor{camel!60}{$\bullet$}) and reference-based (\textcolor{frenchblue!50}{$\bullet$}) evaluation.}
\label{tab:results_word_level}
\end{table}

\begin{table}[t]
\footnotesize
\centering
\begin{tabular}{ccccc}
\toprule
\textsc{Score} & \texttt{zh-en} & \texttt{en-de} & \texttt{en-ru} & \texttt{All}\\ \midrule
\multicolumn{1}{l}{$\hat{y}_{\textsc{src}}$} & 0.73 & 0.75 & 0.79 & 0.78 \\
\multicolumn{1}{l}{$\hat{y}_{\textsc{ref}}$} & 0.75 & 0.74 & 0.75 & 0.77 \\
\multicolumn{1}{l}{$\hat{y}_{\textsc{src+ref}}$} & 0.78 & 0.79 & 0.82 & 0.82 \\
\multicolumn{1}{l}{$\hat{y}_{\textsc{sl}}$$^\dagger$}  & 0.90 & 0.92  & 0.92 & 0.92 \\\bottomrule
\end{tabular}
\caption{Pearson correlations between the regression scores produced by \textbf{x\textsc{comet-xxl}} ($\hat{y}_{\textsc{src}}$, $\hat{y}_{\textsc{ref}}$, $\hat{y}_{\textsc{src+ref}}$, $\hat{y}_{\textsc{sl}}$) and the MQM inferred score, $\hat{y}_{\textsc{mqm}}$, computed from the identified error spans. $^\dagger$The computation of $\hat{y}_{\textsc{sl}}$, contrary to the computation of the other regression scores, makes direct use of $\hat{y}_{\textsc{mqm}}$~(see Equation~\ref{eq:y_sl_computation}).}
\label{tab:corr_features} 
\end{table}

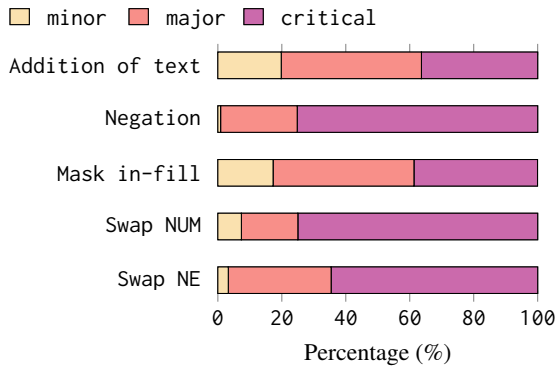
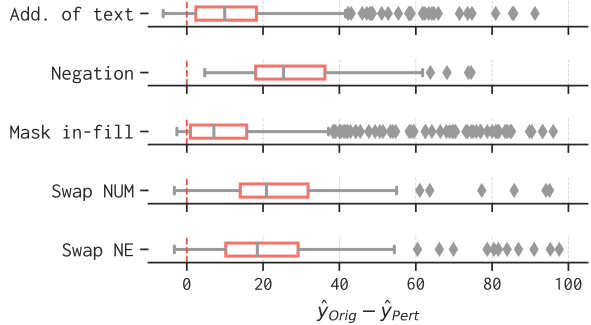
\begin{figure*}[t]
\begin{subfigure}[t]{0.495\textwidth}
\centering
\definecolor{critical}{HTML}{a90674}
\definecolor{major}{HTML}{f6736b}
\definecolor{minor}{HTML}{fae1af}
\definecolor{noerror}{HTML}{d2e8f1}

\pgfplotsset{width=6cm, height=5cm,
    /pgfplots/ybar legend/.style={
    /pgfplots/legend image code/.code={%
      \draw[##1,/tikz/.cd,yshift=-0.25em]
        (0cm,0cm) rectangle (7pt,0.8em);},
  },}
    \flushleft
    \small
    \begin{tikzpicture}  
    \begin{groupplot}[
          group style={
          group name=plot,
          horizontal sep=0pt,
          vertical sep=0pt,
          group size=5 by 1},]
      \nextgroupplot[
            xbar stacked,
            xmin=-0.5, xmax=20.5,
            bar width=10pt,
            ytick={0, 1, 2, 3, 4},
            yticklabels={
            \texttt{Swap NE},
            \texttt{Swap NUM}, 
            \texttt{Mask in-fill},
            \texttt{Negation},
            \texttt{Addition of text},
            },
            xtick={0, 4, 8, 12, 16, 20},
            xticklabels={\texttt{0}, \texttt{20}, \texttt{40}, \texttt{60}, \texttt{80}, \texttt{100}},
            ytick style={draw=none},
            ylabel style={align=left,
                        width=0pt,
                        },
            axis line style={draw=none},
            legend cell align=left,
            legend style={
                /tikz/column 2/.style={column sep=6pt},
                at={(current bounding box.west)}, 
                anchor=west,
                column sep=1.1ex,
                font=\small,
                draw=none,
                yshift=2.25cm,
            },
            legend columns=-1,
            ]  

        % \addplot[xbar, fill=noerror,  postaction={}] coordinates {
        %     (0.65, 4) %
        %     (0.01, 3) %
        %     (1.38, 2) %
        %     (0.50, 1) %
        %     (0.58, 0) %
        % };
        \addplot[xbar, fill=minor,  postaction={}] coordinates {
            (3.96, 4) %
            (0.17, 3) %
            (3.45, 2) %
            (1.47, 1) %
            (0.65, 0) %
        };
        \addplot[xbar, fill=major!80,  postaction={}] coordinates {
            (8.77, 4) %
            (4.79, 3) %
            (8.81, 2) %
            (3.53, 1) %
            (6.43, 0) %
        };
        \addplot[xbar, fill=critical!60,  postaction={}] coordinates {
            (7.27, 4) %
            (15.04, 3) %
            (7.73, 2) %
            (15, 1) %
            (12.92, 0) %
        };

        \legend{
            % \texttt{no error},
            \texttt{minor},
            \texttt{major},
            \texttt{critical}
            }
        ]
        ]
    \end{groupplot}
    % Manually added nodes        
    \node[below=0.45cm of plot c1r1.south, anchor=north] {\small{Percentage (\%)}}; 
    \end{tikzpicture} 
\caption{Percent of error types on data with critical errors (for both \texttt{zh-en} and \texttt{he-en} data), as predicted by \textbf{x\textsc{comet-xxl}}.}
\label{fig:error_type_smaug_distribution}
\end{subfigure} \quad 
\begin{subfigure}[t]{0.495\textwidth}
\centering
\includegraphics[width=\linewidth]{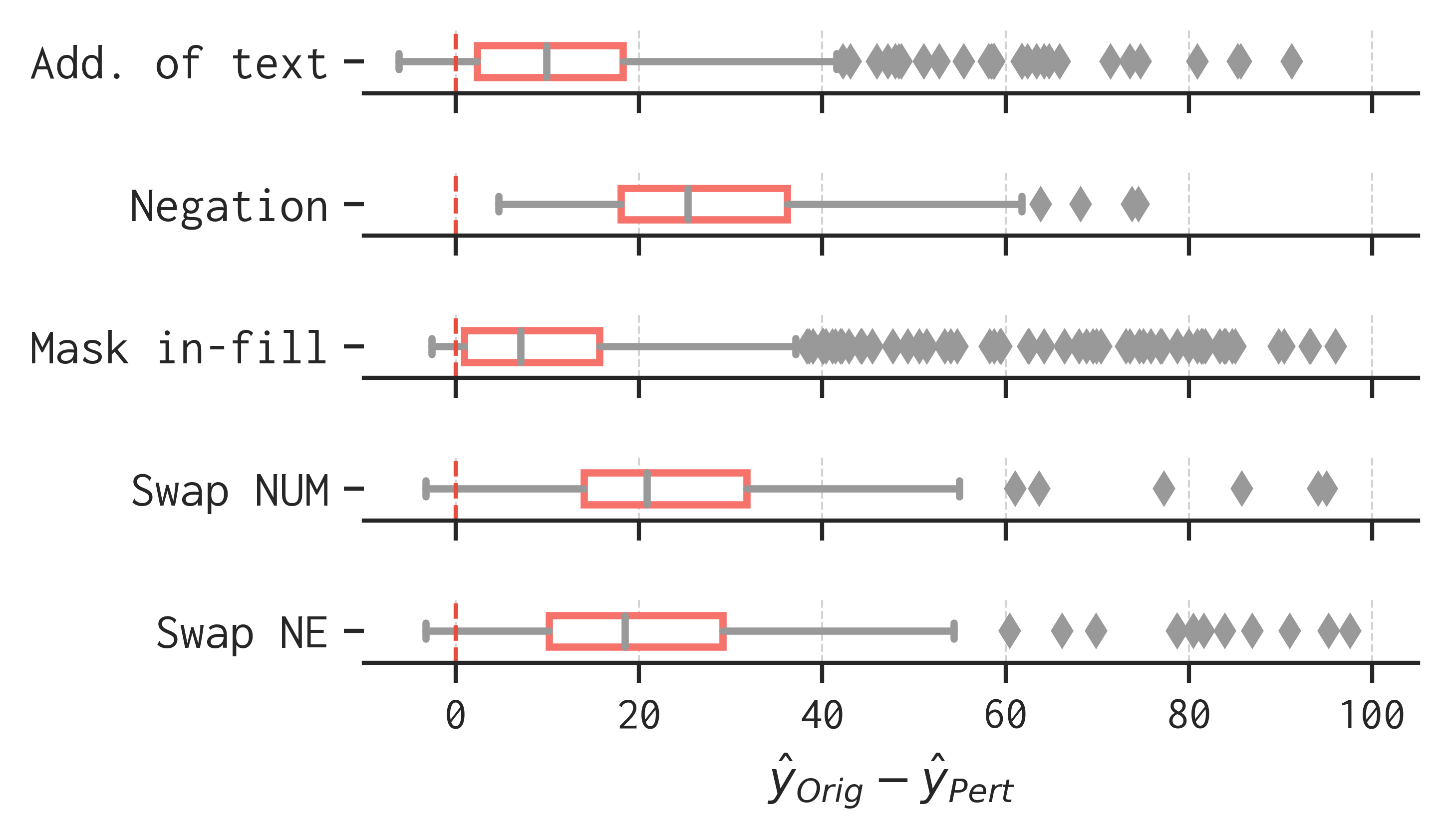}
\caption{Impact of the perturbations, as measured by the difference in \textbf{x\textsc{comet-xxl}} ($\hat{y} = \hat{y}_\textsc{sl}$) between the original and the perturbed translation, on the \texttt{zh-en} data.}
\label{fig:smaug_sensitivity_scores}
\end{subfigure}
\caption{Analysis of x\textsc{comet-xxl} for data with localized critical errors in terms of (a) distribution of error severities for the predicted error spans, and (b) sensitivity of the sentence-level scores.}
\label{fig:smaug_figure}
\end{figure*}

\paragraph{Interplay of error spans and sentence-level scores.} Table \ref{tab:corr_features} shows a strong correlation between the different score types predicted by x\textsc{comet} and the MQM inferred score derived exclusively from error spans. This interplay is highly important: the predicted error spans may be valuable, not just for the sake of accuracy but also for interpretability. Interestingly, these high correlations with the predicted scores from each forward pass~($\hat{y}_{\textsc{src}}$, $\hat{y}_{\textsc{ref}}$, $\hat{y}_{\textsc{src+ref}}$) are obtained despite no explicit alignment mechanism governing the relationship between the predictions of the sentence-level and word-level heads. We hypothesize that it is thus the shared encoder that, during the multi-task training, aligns the representations between the two tasks. As such, x\textsc{comet} provides, through its predicted error spans, a potential lens through which we can better understand, contextualize, and even debug its own sentence-level predictions.

\section{Robustness of x\textsc{comet} to pathological translations}
In the previous section, we have shown that x\textsc{comet} metrics exhibit state-of-the-art correlations with human judgements when evaluating on high-quality MQM annotations. However, more often than not, these MQM annotations are highly unbalanced and contain little to no major or critical errors. As such, they may not offer a full picture of the metrics' performance. In this section, we shift our focus to studying how x\textsc{comet} metrics behave when evaluating translations with localized major or critical errors, such as named-entity errors or mismatches in numbers, and highly pathological translations, such as hallucinations. 

\subsection{Localized errors}
\label{sec:localized-errors}
We employ SMAUG~\cite{alves-etal-2022-robust}\footnote{\url{https://github.com/Unbabel/smaug}}, a tool designed to generate synthetic data for stress-testing metrics, to create corrupted translations that contain major or critical errors. Concretely, we generate translations with the following pathologies: addition of text, negation errors, mask in-filling, named entity errors, and errors in numbers. For this evaluation, we use data from the WMT 2023 Metrics shared task.\footnote{At the time of writing, the human MQM annotations for this data have not been released. Nevertheless, this is not prohibitive for conducting the analysis in this section.} Specifically, we corrupt the released \textit{synthetic} references for which the metrics found no errors.\footnote{This allows us to isolate the effect of the perturbations. In case there are predicted error spans for the transformed translations, these are a result of the perturbation induced.} Moreover, as the full suite of SMAUG transformations can only be applied to English data, we focus solely on Chinese$\to$English (\texttt{zh-en}) and Hebrew$\to$English (\texttt{he-en}) translations. Full details about the corrupted data and examples are shown in Appendix~\ref{app:smaug}. 

\paragraph{x\textsc{comet} predicts most localized errors as major or critical errors.} Table~\ref{tab:percentage-of-noerrors-by-perturb} shows that x\textsc{comet} metrics identify the vast majority of localized errors, with trends varying across scale and language pair. Generally, negation errors and mismatches in numbers are the most easily identified by the metrics. This is interesting: localized errors, such as mismatches in numbers and named-entity errors, had been pinpointed as weaknesses of previous COMET metrics~\citep{amrhein-sennrich-2022-identifying, raunak-etal-2022-salted}. This earlier limitation seems to now have been addressed successfully. In fact, the results in Figure~\ref{fig:error_type_smaug_distribution} show that most of these errors are indeed predicted as critical errors. One plausible hypothesis for these improvements is the incorporation of datasets that contain several negative translations, such as DEMETR, MLQE-PE, and synthetic hallucinations into our training set.
 
\paragraph{x\textsc{comet} sentence-level scores are sensitive to localized perturbations.} Figure~\ref{fig:smaug_sensitivity_scores}\footnote{Results for \texttt{he-en} and x\textsc{comet-xl} can be found in Appendix~\ref{app:smaug}.} shows that localized errors can lead to significant decreases in the predicted sentence-level scores, with perturbation-wise trends mirroring those of the error span predictions: the most pronounced decreases are found for negation errors and mismatches in numbers and named-entities~(median decreases of around 20 points). The distribution of the decreases in quality also reveals two relevant trends: (i)~localized perturbations can cause x\textsc{comet-xxl} to shift from a score of a perfect translation to that of an unrelated translation, and (ii)~the behavior of x\textsc{comet-xxl} is not perfect and can be further improved: in some rare cases, perturbations may actually lead increase in the score. Nevertheless, upon closer inspection, we found that, for over 90\% of these cases, the decrease is smaller than 1 point.

\begin{table}[t]
\centering
\footnotesize
\begin{tabular}{lrrrrr}
\toprule
   & \multicolumn{2}{c}{\texttt{zh-en}} & & \multicolumn{2}{c}{\texttt{he-en}} \\\cmidrule{2-3}\cmidrule{5-6}
   \textsc{Error}  & \multicolumn{1}{c}{\textsc{xl}} & \multicolumn{1}{c}{\textsc{xxl}}  & & \multicolumn{1}{c}{\textsc{xl}}  & \multicolumn{1}{c}{\textsc{xxl}} \\ \midrule
   \texttt{Add. of text} & 3.66 & 10.7 & & 6.15 & 7.35\\
   \texttt{Negation} & 0.20 & 0.20& & 3.89 & 4.90\\
   \texttt{Mask in-fill} & 5.01 & 17.0& & 4.78& 3.92\\
   \texttt{Swap NUM} & 3.19 & 2.88& & 0.16& 0.00\\
   \texttt{Swap NE} & 3.66 & 6.94& & 9.81& 7.01\\\cdashlinelr{1-6}
   \texttt{All} & \textbf{2.24} & 10.7 & & 9.81 & \textbf{7.00} \\\bottomrule
\end{tabular}
\caption{Percentage ($\%$) of translations, segmented by perturbation type, that are predicted to have no errors~($\downarrow$). We show results for both \texttt{zh-en} and \texttt{he-en} language pairs across x\textsc{comet} (\textsc{xl} and \textsc{xxl}) sizes.}
\label{tab:percentage-of-noerrors-by-perturb}
\end{table}

\subsection{Hallucinations}
Hallucinations lie at the extreme-end of machine translation pathologies~\citep{raunak-etal-2021-curious}, and can have devastating impact when models are deployed \textit{in the wild}. Yet, these translations are often overlooked when assessing the performance of different translation systems. Their rarity means that performance, usually judged according to an aggregated corpus-level score, may remain largely unperturbed by a very small number of hallucinations.\footnote{For example, hallucination rates with state-of-the-art translation systems on high-resource language pairs are extremely rare: \citet{raunak-etal-2022-salted} found only five hallucinations in over 100k translations when using Microsoft's translation system by way of their paid public APIs; and, \citet{guerreiro-etal-2023-hallucinations} found zero hallucinations with a 12B M2M model~\citep{fan-etal-2021-m2m} when tested on more than 35k translations. This trend, however, was not found for medium and low-resource translation directions, where hallucination rates can be well above 10\% for some language pairs.} In this section, we want to assess how the x\textsc{comet} metrics rank hallucinations among other translations. To do so, we will use the German$\to$English hallucination benchmark introduced in~\citet{guerreiro-etal-2023-looking}. This benchmark involves over 3.4k translations of different error types, including omissions, named-entity errors, and hallucinations~(oscillatory, fully, and strongly detached). For a metric that has not been trained explicitly to rank translations, the benchmark is quite challenging: hallucinations should be ranked below other severe errors and incorrect translations. We provide examples of the hallucinations in the dataset in Appendix~\ref{app:hallucinations}.

\begin{table}[t]
    \centering
    % First minipage for the table
        \footnotesize
        \begin{tabular}{lccc}
        \toprule
        \textsc{Metric} & \texttt{All} & \texttt{Full Det.} & \texttt{Osc.} \\ \midrule
        \textcolor{frenchblue!50}{$\bullet$} \textsc{Bleurt-20} & 0.824 & 0.892 & 0.799 \\
        \textcolor{frenchblue!50}{$\bullet$} \textsc{Comet-22} & 0.829 & 0.878 & 0.883\\ 
        \textcolor{camel!60}{$\bullet$} \textsc{CometKiwi-xxl} & 0.839 & 0.834 & 0.902\\ \midrule
        \textcolor{frenchblue!50}{$\bullet$} \textbf{x\textsc{comet-xl}} & 0.865 & 0.907 & 0.922  \\
        \textcolor{frenchblue!50}{$\bullet$} \textbf{x\textsc{comet-xxl}} & 0.890 & \textbf{0.964} & 0.844 \\\cdashlinelr{1-4}
        \multicolumn{4}{c}{\textit{QE scores from the error spans ($\hat{y} = \hat{y}_{\textsc{src}}$)}} \\
        \textcolor{camel!60}{$\bullet$} \textbf{x\textsc{comet-xl}} (\textsc{src}) & 0.885 & 0.924 & \textbf{0.944} \\
        \textcolor{camel!60}{$\bullet$} \textbf{x\textsc{comet-xxl}} (\textsc{src}) & \textbf{0.902} & 0.959 & 0.866 \\\bottomrule
        \end{tabular}
        % \begin{tabular}{lcc}
        % \toprule
        % \textsc{Metric} & \texttt{All} & \texttt{Fully Detached}\\ \midrule
        % \textcolor{frenchblue!50}{$\bullet$} \textsc{Bleurt-20} & 0.824 & 0.892  \\
        % \textcolor{frenchblue!50}{$\bullet$} \textsc{Comet-22} & 0.829 & 0.878 \\ 
        % \textcolor{camel!60}{$\bullet$} \textsc{CometKiwi-xxl} & 0.839 & 0.834\\ \midrule
        % \textcolor{frenchblue!50}{$\bullet$} \textbf{x\textsc{comet-xl}} & 0.865 & 0.907   \\
        % \textcolor{frenchblue!50}{$\bullet$} \textbf{x\textsc{comet-xxl}} & 0.890 & \textbf{0.964}  \\\cdashlinelr{1-3}
        % \multicolumn{3}{c}{\textit{QE scores from the error spans ($\hat{y} = \hat{y}_{\textsc{src}}$)}} \\
        % \textcolor{camel!60}{$\bullet$} \textbf{x\textsc{comet-xl}} (\textsc{src}) & 0.885 & 0.924  \\
        % \textcolor{camel!60}{$\bullet$} \textbf{x\textsc{comet-xxl}} (\textsc{src}) & \textbf{0.902} & 0.959 \\\bottomrule
        % \end{tabular}
        \captionof{table}{Hallucination detection performance on the \texttt{de-en} hallucination benchmark from~\citet{guerreiro-etal-2023-looking} as measured by AUROC ($\uparrow$) for reference-free (\textcolor{camel!60}{$\bullet$}) and reference-based (\textcolor{frenchblue!50}{$\bullet$}) quality metrics. We report results for all the dataset, for fully detached, and oscillatory hallucinations separately.}
        \label{tab:hallucination_detection}
\end{table}

\begin{figure}[t]
        \centering
        \includegraphics[width=\linewidth]{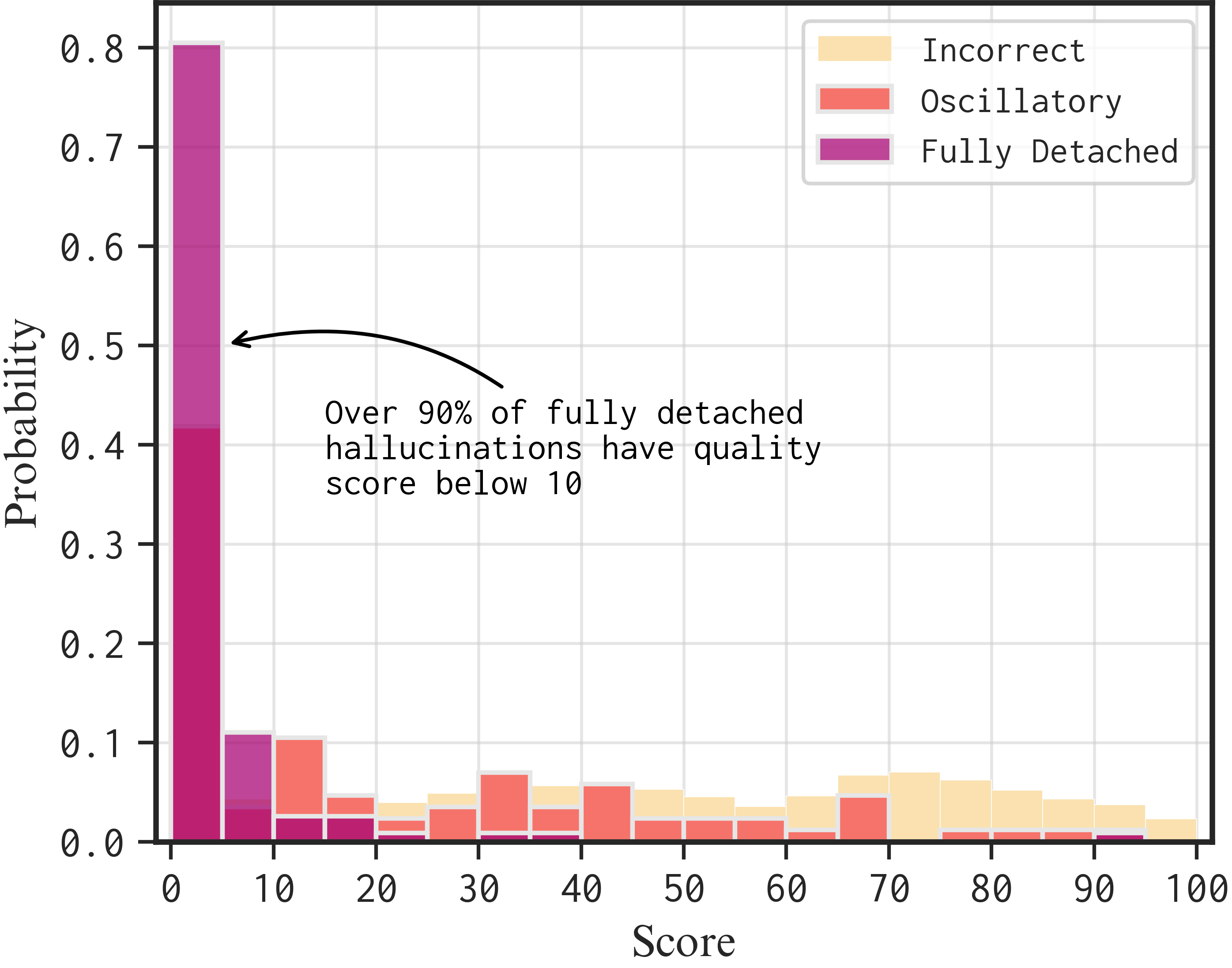}
        \caption{Category-wise distribution of \textbf{x\textsc{comet-xxl}} scores on the hallucination benchmark.}
        \label{fig:hallucination_distribution}
\end{figure}

\paragraph{x\textsc{comet} metrics can distinguish hallucinations from other translations.} The results in Table~\ref{tab:hallucination_detection} show that both x\textsc{comet} metrics largely rank hallucinations lower than other errors. This is especially true for the most severe type of hallucination~(fully detached), for which the AUROC exceeds 95 for the \textsc{xxl} metric. In fact, Figure~\ref{fig:hallucination_distribution} reveals that x\textsc{comet-xxl} assigns over 90\% of these fully detached hallucinations a score under 10. Relative to previous metrics, x\textsc{comet} achieves overall improvements. Interestingly, we also find that \textsc{src}-based evaluation (i.e., without the use of a reference translation) can reap benefits in this scenario. We hypothesize that this is due to the metric over-relying on the reference when it is made available~\citep{rei-etal-2023-inside}. While hallucinations contain content that is detached from the source, some of their text may still overlap (even if just lexically) with the reference text~(e.g., in strongly detached or oscillatory hallucinations), leading to higher scores. In future work, it would be interesting to explore whether these trends hold for other language pairs, including low-resource ones, through the use of multilingual hallucination benchmarks like HalOmi~\citep{dale2023halomi}.\footnote{The HalOmi benchmark is yet to be publicly released.}

\section{Conclusions}
We introduced x\textsc{comet}: a novel suite of metrics for machine translation evaluation that effectively combines sentence-level prediction with fine-grained error span prediction. Through extensive experiments, we have shown that x\textsc{comet} is a state-of-the-art metric at all relevant vectors of evaluation: sentence-level, system-level, and error span prediction. Notably, through x\textsc{comet}'s capabilities to predict error spans, we can not only obtain useful signals for downstream prediction (either directly through error span prediction or by informing sentence-level scores) but also gain access to a lens through which we can better understand and interpret its predictions. Finally, we also stress-tested the suite of metrics by analyzing their behavior on scoring localized critical errors and hallucinations: x\textsc{comet} metrics identify the vast majority of localized errors and can appropriately penalize the severity of hallucinations.

We hope x\textsc{comet} can serve as a further step towards more detailed and informed machine translation evaluation. The full suite of metrics (x\textsc{comet-xl} and x\textsc{comet-xxl}) will be made available through the HuggingFace Hub: \href{https://huggingface.co/Unbabel}{\texttt{https://huggingface.co/Unbabel}}.

% \section*{Limitations}

\section*{Acknowledgements}
We are grateful to José Pombal, José G. C. de Souza and Sweta Agrawal for their valuable feedback and discussions.

This work was supported by the Portuguese Recovery and Resilience Plan (PRR) through project C645008882-00000055, Center for Responsible AI, by the European Research Council (DECOLLAGE, ERC-2022-CoG 101088763), by EU’s Horizon Europe Research and Innovation Actions (UTTER, contract 101070631), and by the Funda\c{c}\~{a}o para a Ci\^{e}ncia e Tecnologia (contracts UIDB/50021/2020 and UIDB/50008/2020). We also thank the HPC resources from GENCI-IDRIS (Grants 2023--AD011014714, 2023--AD0110146A68R1 and AD011012377R2).

% Entries for the entire Anthology, followed by custom entries
\bibliography{anthology,custom}
\bibliographystyle{acl_natbib}

\clearpage

\appendix 
\onecolumn
\begin{center}
    \Large{\textbf{Supplemental Material}}
\end{center}

% Define custom YAML highlighting
\lstdefinestyle{yamlStyle}{
  basicstyle=\ttfamily\small,
  breaklines=true,
  frame=tb,
  columns=fullflexible,
  commentstyle=\color{gray},
  morekeywords={true,false},
  keywordstyle=\color{blue},
  stringstyle=\color{red},
  identifierstyle=\color{black},
  morestring=[b]',
  morestring=[b]",
  morecomment=[s]{\#}{\ },
  morekeywords={unified_metric,class_path,init_args,nr_frozen_epochs,keep_embeddings_frozen,optimizer,encoder_learning_rate,learning_rate,layerwise_decay,encoder_model,pretrained_model,sent_layer,word_layer,layer_transformation,layer_norm,loss,dropout,batch_size,hidden_sizes,activations,input_segments,word_level_training,loss_lambda,error_labels,cross_entropy_weights},
}

\section{Examples of synthetic hallucinations in the training data}
\label{app_sec:synthetic_examples_halls}
We show in Table~\ref{tab:synthetic-hallucination-examples} one example for each type of synthetic hallucination that we use to augment x\textsc{comet}'s training data.

\begin{table*}[h!]
\small
\begin{tabular}{p{6in}}
\toprule
\textbf{Source:}  \\
Touristen in Portugal in Panik versetzt, nachdem ein tieffliegender Militärjet Strand überfliegt \\\cdashlinelr{1-1}
\textbf{Reference:}  \\
"Tourists in Portugal are left terrified as a low-flying military jet flies skims beach"\\\cdashlinelr{1-1}
\textbf{Detached Hallucination (random):}  \\
\hlmajortab{And best with a deal. I am cautiously optimistic that this will work.}\\\cdashlinelr{1-1}
\\
\textbf{Source:}  \\
Komet entdeckt: Interstellarer Gast kreuzt durch unser Sonnensystem \\\cdashlinelr{1-1}
\textbf{Reference:}  \\
Comet discovered: An interstellar guest crosses through our solar system\\\cdashlinelr{1-1}
\textbf{Detached Hallucination (semantically similar):}  \\
\hlmajortab{Comet crossed by other star solar system} \\\cdashlinelr{1-1}
\\
\textbf{Source:}  \\
Wie ist jetzt die Situation auf der Insel? \\\cdashlinelr{1-1}
\textbf{Reference:}  \\
What is the situation on the island now?\\\cdashlinelr{1-1}
\textbf{Oscillatory Hallucination:}  \\
What is the situation on \hlmajortab{the island the island the island the island the island the island the island} now?\\
\bottomrule
\end{tabular}
\caption{Examples of synthetically-generated hallucinations for Phases II and III of x\textsc{comet}'s training.}
\label{tab:synthetic-hallucination-examples}
\end{table*}

\section{Hyperparameters}
\label{app_sec:hyperparameters}

\subsection{Setting the parameter $\lambda$ for curriculum training}
To obtain $\lambda$ for each Phase, we do hyperparameter tuning with Optuna~\cite{optuna_2019}, running over 20 trials changing $\lambda$ in adequate intervals depending on the objective of the phase (e.g., values closer to 1 for Phase II, and values closer to 0 for Phase III).

\subsection{Hyperparamers used for each phase}
You can find the hyperparameters used to train models from Phase I to Phase III in Listings~\ref{lst.phaseI} and~\ref{lst.phaseIIandIII}.\footnote{We omit the config arguments that are respective to the encoder model, such as \texttt{word\_layer} and \texttt{hidden\_sizes}.} Note that the only difference between the two phases, apart from the training data (which we describe in~\ref{ssec:corpora}), is the value of the $\lambda$ parameter. Regarding the class weights $\bm{\alpha}$, we have also optimized this parameter using Optuna~\cite{optuna_2019}, and keep them fixed throughout Phases II and III. As expected, the \textit{optimal} weights reflect the class-inbalance, assigning the smallest weight to the \textsc{ok} tag and the largest to the \textsc{maj} and \textsc{crit} tags.

\begin{lstlisting}[basicstyle=\tiny,style=yamlStyle, caption=Relevant hyperparameters used for training Phase - I models (Section \ref{ssec:architecture}) using \textsc{Comet} framework., float, label=lst.phaseI]
class_path: UnifiedMetric
init_args:
nr_frozen_epochs: 0.3
keep_embeddings_frozen: True
optimizer: AdamW
encoder_learning_rate: 1.83e-06
learning_rate: 3.66e-06
layerwise_decay: 0.983
sent_layer: mix
layer_transformation: sparsemax
layer_norm: False
loss: mse
dropout: 0.1
batch_size: 32
activations: Tanh
input_segments:
  - mt
  - src
  - ref
word_level_training: False 
\end{lstlisting}
\begin{small}
\begin{lstlisting}[style=yamlStyle, caption=Relevant hyperparameters used for training Phase II and III models (Section \ref{ssec:architecture}) using \textsc{Comet} framework. Note that the only difference between the two phases is the loss $\lambda$ parameter (Eq. \ref{eq:comb_loss})., float, label=lst.phaseIIandIII]
class_path: UnifiedMetric
init_args:
nr_frozen_epochs: 0.3
keep_embeddings_frozen: True
optimizer: AdamW
encoder_learning_rate: 1.0e-06
learning_rate: 3.66e-06
layerwise_decay: 0.983
sent_layer: mix
layer_transformation: sparsemax
layer_norm: False
loss: mse
dropout: 0.1
batch_size: 32
activations: Tanh
input_segments:
  - mt
  - src
  - ref
word_level_training: true
loss_lambda: 0.983 (II) / 0.055 (III)
error_labels:
  - minor
  - major
  - critical
cross_entropy_weights:
  - 0.08
  - 0.486
  - 0.505
  - 0.533
\end{lstlisting}
\end{small}

\section{SMAUG data}
\label{app:smaug}
\subsection{Data construction}
In Section \ref{sec:localized-errors} we analysed how x\textsc{Comet} behaves when the translation hypothesis contains localized major/critical errors using synthetic data created with SMAUG~\cite{alves-etal-2022-robust}. The analysed pathologies are: hallucinations via addition of text, negation errors, hallucinations via mask in-fill, swap of numbers and swap of named entities.\footnote{On the hallucinations via addition of text and via mask in-fill, SMAUG encourages to add tokens or replace masked tokens with other tokens that preserve fluency. As such, there may be some perturbed translations that are close to paraphrases of the original translation. We hypothesize this is the reason why there is a bigger percentage of minor errors when compared to other error types~(e.g., negation errors).} Table~\ref{tab:smaug} presents a summary of the examples created using SMAUG. We also provide examples of each error category in~\ref{tab:smaug-examples}.

\subsection{Additional results}
\begin{figure}[h]
    % Top row of subfigures
    \begin{subfigure}[t]{0.495\textwidth}
        \centering
        \includegraphics[width=\linewidth]{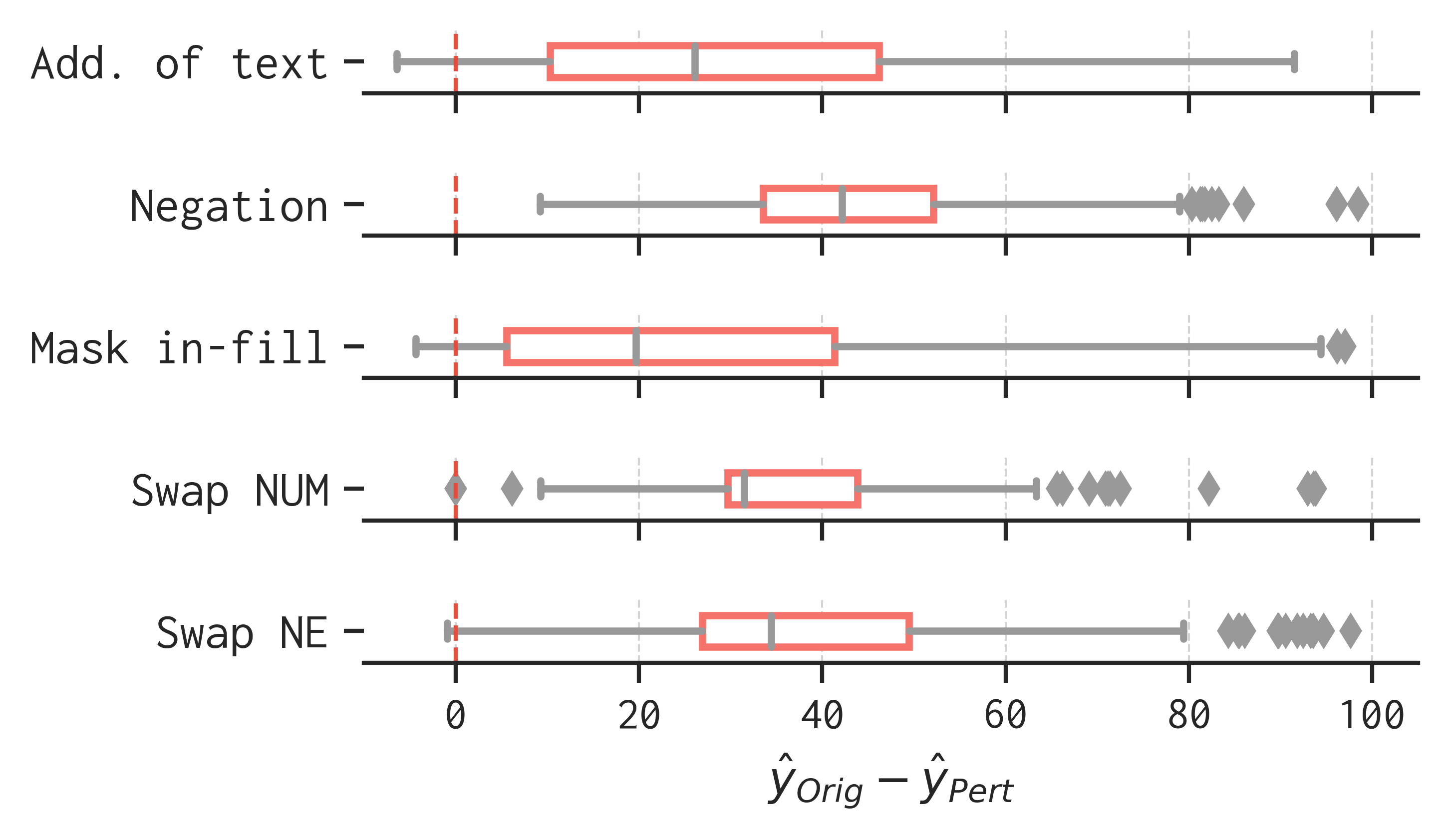}
        \caption{\textbf{x\textsc{comet-xxl}} ($\hat{y} = \hat{y}_\textsc{sl}$) on \texttt{he-en} data.}
        \label{fig:smaug_xxl_zhen}
    \end{subfigure} 
    \hfill % ensures that they are spaced out to the margins
    \begin{subfigure}[t]{0.495\textwidth}
        \centering
        \includegraphics[width=\linewidth]{figs/smaug_all_models_boz.png}
        \caption{\textbf{x\textsc{comet-xxl}} ($\hat{y} = \hat{y}_\textsc{sl}$) on \texttt{zh-en} data.}
        \label{fig:smaug_xxl_zhen}
    \end{subfigure}

    \vspace{1em} % provides some space between the rows

    % Bottom row of subfigures
    \begin{subfigure}[t]{0.495\textwidth}
        \centering
        \includegraphics[width=\linewidth]{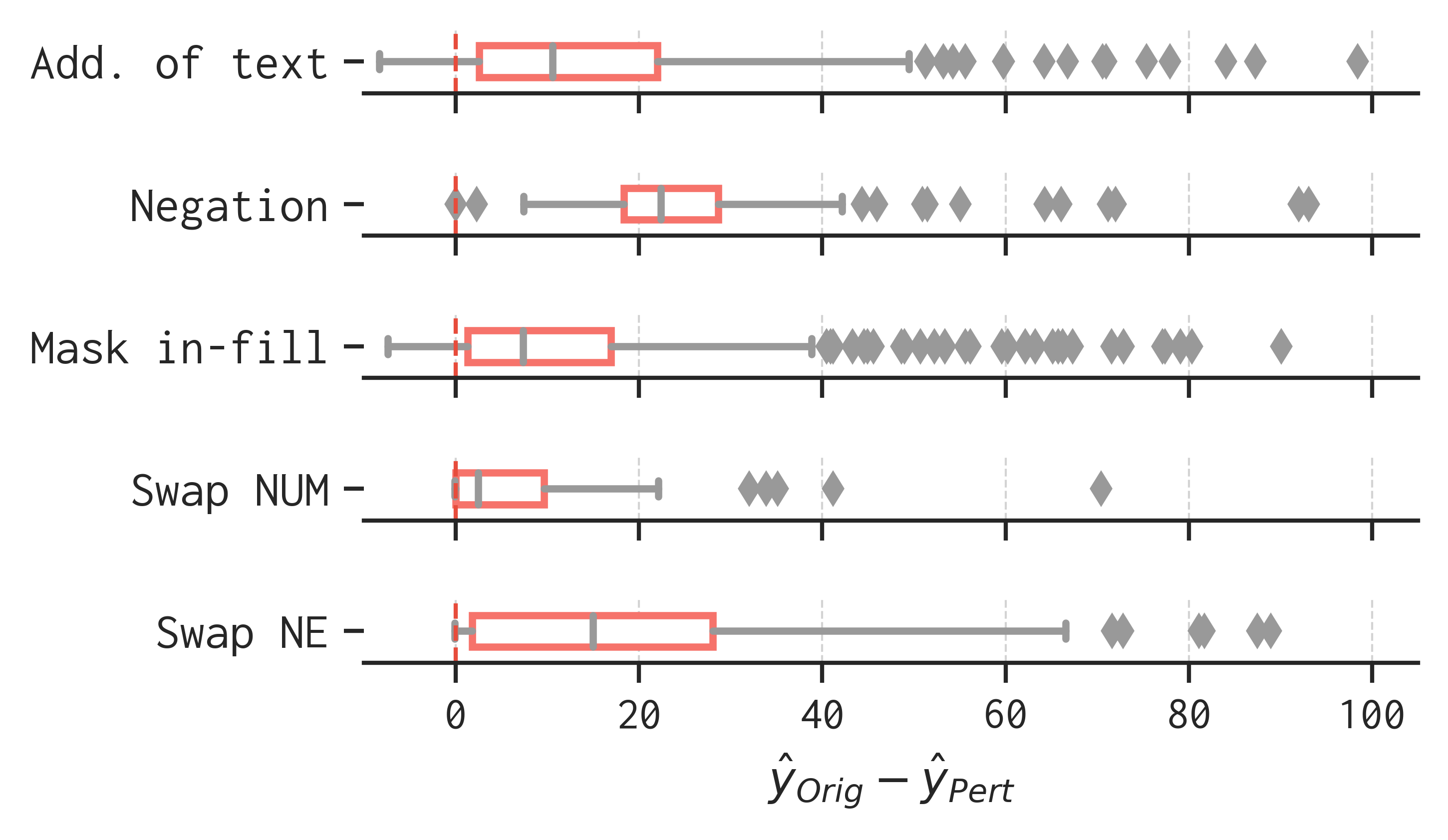}
        \caption{\textbf{x\textsc{comet-xl}} ($\hat{y} = \hat{y}_\textsc{sl}$) on \texttt{he-en} data.}
        \label{fig:smaug_xxl_zhen}
    \end{subfigure} 
    \hfill % ensures that they are spaced out to the margins
    \begin{subfigure}[t]{0.495\textwidth}
        \centering
        \includegraphics[width=\linewidth]{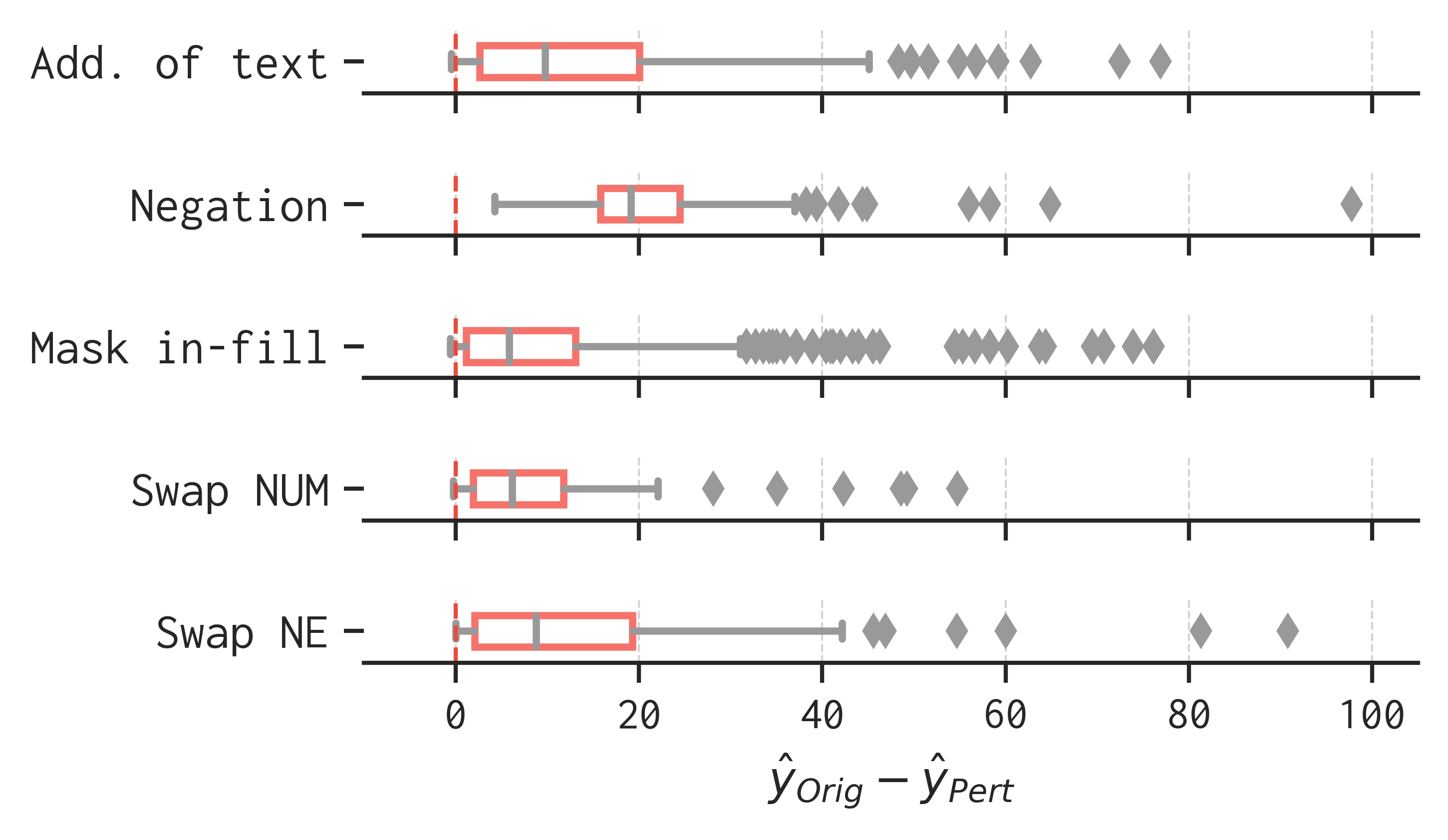}
        \caption{\textbf{x\textsc{comet-xl}} ($\hat{y} = \hat{y}_\textsc{sl}$) on \texttt{zh-en} data.}
        \label{fig:smaug_xxl_zhen}
    \end{subfigure}

    \caption{Impact of the perturbations, as measured by the difference in \textbf{x\textsc{comet-xxl}} ($\hat{y} = \hat{y}_\textsc{sl}$) between the original and the perturbed translation with both models.}
    \label{fig:smaug_figure_impact}
\end{figure}

We provide additional results on the impact of SMAUG perturbations for \texttt{zh-en} and \texttt{he-en} data for both the \textsc{xl} and \textsc{xxl} x\textsc{comet} models in Figure~\ref{fig:smaug_figure_impact}.

\begin{table}[t]
\small
\centering
%\resizebox{\columnwidth}{!}{
\begin{tabular}{lrrr}
\toprule
   \textsc{Error} & \texttt{zh-en} & \texttt{he-en} \\\midrule
   \texttt{Add. of text} & 1516 & 1490\\
   \texttt{Negation} & 498 &  637\\
   \texttt{Mask in-fill} & 1716 & 1659\\
   \texttt{Swap NUM} & 313 & 214 \\
   \texttt{Swap NE} & 519 & 586 \\\midrule
   \texttt{Total} & 4762 &  4586\\
\bottomrule
\end{tabular}
\caption{Number of examples for each category, synthetically-created using SMAUG~\cite{alves-etal-2022-robust} for \texttt{zh-en} and \texttt{he-en}}  
\label{tab:smaug}
%}
\end{table}

\begin{table}[h!]
\small
\begin{tabular}{p{6in}}
\toprule
\textbf{Source:}  \\
\zh{“这无疑是火上浇油，支持了对华强硬派的观点”。} \\\cdashlinelr{1-1}
\textbf{Translation:}  \\
"This undoubtedly added fuel to the fire, supporting the views of the hardliners towards China." \\\cdashlinelr{1-1}
\textbf{Reference:}  \\
"This undoubtedly adds fuel to the fire and supports the views of the hardliners towards China."\\\cdashlinelr{1-1}
\textbf{\texttt{Add. of text}:}  \\
\hlmajortab{he said the media was not "responsible" for such behavior. }"This undoubtedly added fuels to the fire, thereby supporting the views of the hardliners towards China." \\\cdashlinelr{1-1}
\\
\textbf{Source:}  \\
\zh{请注意本文件的某些内容可能涉及专利。} \\\cdashlinelr{1-1}
\textbf{Translation:}  \\
Please be aware that this text may include references to patents in some places. \\\cdashlinelr{1-1}
\textbf{Reference:}  \\
Please be aware that this text may include references to patents in some places.\\\cdashlinelr{1-1}
\textbf{\texttt{Negation}:}  \\
Please be aware that this text \hlmajortab{does not} include references to patents in some places. \\\cdashlinelr{1-1}
\\
\textbf{Source:}  \\
\zh{任何本文件适用的产品均应满足实现级别 1 的要求。} \\\cdashlinelr{1-1}
\textbf{Translation:}  \\
Any product applicable to this document should meet the requirements of implementation level 1. \\\cdashlinelr{1-1}
\textbf{Reference:}  \\
Each item to which this record relates must follow to the Execution Level 1 details.\\\cdashlinelr{1-1}
\textbf{\texttt{Mask in-fill}:}  \\
Any product applicable to \hlmajortab{the project, which does not} meet the requirements of implementation level 1.\\\cdashlinelr{1-1}
\\
\textbf{Source:}  \\
\zh{算我倒霉，懒得为了个200块的东西被你们折腾。} \\\cdashlinelr{1-1}
\textbf{Translation:}  \\
I'm unfortunate because I'm too lazy to be hurled by you for 200 yuan stuff. \\\cdashlinelr{1-1}
\textbf{Reference:}  \\
I'm unfortunate because I'm too lazy to be hurled by you for 200 yuan stuff.\\\cdashlinelr{1-1}
\textbf{\texttt{Swap NUM}:}  \\
I'm unfortunate because I'm too lazy to be hurled by you for the \hlmajortab{5} yuan stuff.\\\cdashlinelr{1-1}
\\
\textbf{Source:}  \\
\zh{本文件由中华人民共和国工业和信息化部提出并归口。} \\\cdashlinelr{1-1}
\textbf{Translation:}  \\
This document was proposed and managed by the Ministry of Industry and Information Technology of the People's Republic of China.\\\cdashlinelr{1-1}
\textbf{Reference:}  \\
The People's Republic of China's Ministry of Industry and Information Technology is responsible for the proposal and administration of this document.\\\cdashlinelr{1-1}
\textbf{\texttt{Swap NE}:}  \\
This document was proposed and managed by the \hlmajortab{Government} of the People's Republic of China.
\\\cdashlinelr{1-1}
\\
% \cdashlinelr{1-1}
% \\
\bottomrule
\end{tabular}
\caption{Synthetically-generated errors (\hlmajortab{highlighted in pink}) created with SMAUG~\cite{alves-etal-2022-robust} for \texttt{zh-en} to assess whether x\textsc{Comet} can detect localized major and critical errors.}
\label{tab:smaug-examples}
\end{table}

%\begin{table}[t]
%    \centering
%    \small    
%    \begin{tabular}{l ccccc}
%        \toprule
%        \bf Method & \bf zh-en & \bf en-de & \bf he-en  & & \bf avg. \\
%        \midrule
%        Baseline & 0.219 & 0.167 & 0.083 & & 0.156 \\
%        \midrule
%        \textbf{x\textsc{comet-xl}} & 0.199 & 0.195 & 0.139 & & 0.178 \\
%        \textbf{x\textsc{comet-xxl}} & 0.252 & 0.229 & 0.133 & & 0.204 \\\cdashlinelr{1-6}
%        \multicolumn{6}{c}{\textit{Using a pseudo-reference}} \\
%        \textbf{x\textsc{comet-xl}} & 0.224 & 0.212 & 0.140 & & 0.192 \\
%        \textbf{x\textsc{comet-xxl}} & \textbf{0.269} & \textbf{0.249} & \textbf{0.150} & & \textbf{0.223} \\
        %x\textsc{comet-ps-ref} & 0.259 & 0.270 & \bf 0.125 & & 0.218 \\ \cdashlinelr{1-8}
        %\textsc{UnbabelQi} & 0.249 & 0.227 & 0.111 & & 0.196 \\
        %\textsc{GPT4-QE} & \bf 0.273 &	0.265 & 0.121 & & \bf 0.220 \\
%        \bottomrule
%    \end{tabular}
%    \caption{Results for fine-grained error span detection (Task 2). \textcolor{red}{The?}Evaluation metric is $F_1$ score. We represent zero-shot LPs with $\dagger$. The first two systems are constrained while the other two are unconstrained submissions.
%}
%\label{tab:results_task_2}
%\end{table}

\section{Examples of hallucinations from the benchmark of~\citet{guerreiro-etal-2023-looking}}
\label{app:hallucinations}
We provide in Table~\ref{tab:xcomet_hallucinations_deen} examples of hallucinations from the benchmark that we used, alongside the predicted error spans by x\textsc{comet}.

\begin{table*}[h!]
\small
\begin{tabular}{p{6in}}
\toprule
\textbf{Source:}  \\
Touristen in Portugal in Panik versetzt, nachdem ein tieffliegender Militärjet Strand überfliegt \\\cdashlinelr{1-1}
\textbf{Translation:} \\
Is the whole pancreas removed from each pancreas of pancreas? \\\cdashlinelr{1-1}
\textbf{Reference:}  \\
Is the entire pancreas always removed in pancreatic cancer operations?\\\cdashlinelr{1-1}
\textbf{Oscillatory hallucination with x\textsc{comet} error span predictions:}  \\
Is the \hlcrittab{\textbf{\texttt{CRIT}:} whole pancreas removed from each pancreas of pancreas}?\\\cdashlinelr{1-1}
\\
\textbf{Source:}  \\
Das Teilabonnement für international tätige Juristen. \\\cdashlinelr{1-1}
\textbf{Translation:} \\
The sub-sub-sub-sub-scription for international lawyers. \\\cdashlinelr{1-1}
\textbf{Reference:}  \\
Partial subscriptions for internationally active lawyers.\\\cdashlinelr{1-1}
\textbf{Oscillatory hallucination with x\textsc{comet} error span predictions:}  \\
The \hlcrittab{\textbf{\texttt{CRIT}:} sub-sub-sub-sub-scription} for international lawyers.\\\cdashlinelr{1-1}
\\
\textbf{Source:}  \\
Nur der Stuck an der Decke uns das Wandabendmahl konnten nicht restauriert werden.\\\cdashlinelr{1-1}
\textbf{Translation:} \\
The room was very small and the bathroom was very small. \\\cdashlinelr{1-1}
\textbf{Reference:}  \\
Only the stuccos of the ceiling and the wall could not be restored.\\\cdashlinelr{1-1}
\textbf{Fully detached hallucination with x\textsc{comet} error span predictions:}  \\
\hlcrittab{\textbf{\texttt{CRIT}:} The room was very small and the bathroom was very small.}\\\cdashlinelr{1-1}
\\
\textbf{Source:}  \\
Empfehlenswert gleich mit der Zimmerreservierung zu buchen! \\\cdashlinelr{1-1}
\textbf{Translation:} \\
The staff were very friendly and helpful. The room was clean and comfortable. \\\cdashlinelr{1-1}
\textbf{Reference:}  \\
We recommend booking your treatments together with the hotel booking! \\\cdashlinelr{1-1}
\textbf{Fully detached hallucination with x\textsc{comet} error span predictions:}  \\
\hlcrittab{\textbf{\texttt{CRIT}:} The staff were very friendly and helpful. The room was clean and comfortable.} \\
\bottomrule
\end{tabular}
\caption{Examples of predictions of x\textsc{comet-xxl} for the hallucination benchmark of~\citet{guerreiro-etal-2023-looking}.}
\label{tab:xcomet_hallucinations_deen}
\end{table*}

\end{document}